\documentclass[prd,amsmath,amssymb]{revtex4}
\usepackage{graphicx}% Include figure files
\usepackage{dcolumn}% Align table columns on decimal point
\usepackage{bm}% bold math
\usepackage{cases}
\usepackage{mathtools}

\DeclarePairedDelimiter\ket{\lvert}{\rangle}
\DeclarePairedDelimiterX\braket[2]{\langle}{\rangle}{#1\,\delimsize\vert\,\mathopen{}#2}
\usepackage{pgfplots}
\pgfplotsset{compat=1.8}
\usetikzlibrary{arrows.meta}

\begin{document}

\title{Semantic Wave Functions: Exploring Meaning in Large Language Models Through Quantum Formalism}

\author{Timo\hspace*{1mm}Aukusti\hspace*{1mm}Laine \vspace*{0.5cm}}
\email{timo@financialphysicslab.com}

\begin{abstract}
    Large Language Models (LLMs) encode semantic relationships in high-dimensional vector embeddings. 
    This paper explores the analogy between LLM embedding spaces and quantum mechanics, positing that LLMs 
    operate within a quantized semantic space where words and phrases behave as quantum states. To capture 
    nuanced semantic 
    interference effects, we extend the standard real-valued embedding space to the complex domain, drawing 
    parallels to the double-slit experiment. We introduce a "semantic wave function" to formalize this 
    quantum-derived representation and utilize potential landscapes, such as the double-well potential, 
    to model semantic ambiguity. Furthermore, we propose a complex-valued similarity measure that incorporates 
    both magnitude and phase information, enabling a more sensitive comparison of semantic representations. 
    We develop a path integral formalism, based on a nonlinear Schrödinger equation with a gauge 
    field and Mexican hat potential, to model the dynamic evolution of LLM behavior. This interdisciplinary 
    approach offers a new theoretical framework for understanding and potentially manipulating LLMs, with 
    the goal of advancing both artificial and natural language understanding.
\end{abstract}

\maketitle

\section{Introduction}

Large Language Models (LLMs) have emerged as transformative tools in natural language processing, 
demonstrating remarkable capabilities in tasks ranging from text generation and translation to question 
answering and code completion. At the heart of these models lies a sophisticated mechanism for 
representing text: high-dimensional vector embeddings. These embeddings map words, phrases, and even entire 
documents into a continuous semantic space, where geometric relationships reflect semantic similarities. 
For instance, words with related meanings are positioned closer together, while dissimilar concepts are further apart.

While these embedding spaces are often treated as continuous for practical purposes, a fundamental 
aspect of LLMs hints at an underlying discreteness: their reliance on a finite vocabulary of tokens. 
This discrete foundation suggests that the seemingly continuous semantic space might, in fact, 
possess a quantized structure, analogous to the discrete energy levels observed in quantum systems. 
This inherent quantization prompts a compelling question: can we leverage the powerful theoretical 
frameworks of mathematical physics and tools of quantum mechanics to gain a deeper understanding of the organization 
and dynamics of these semantic spaces? 
Furthermore, if this quantization is valid, could quantum computing, for example, offer new approaches to training or 
exploiting these models, potentially unlocking significant performance gains?

This article embarks on an exploration of this intriguing analogy between LLM embedding spaces and quantum mechanics. 
We focus on the concept of quantization, the wave-like behavior of semantic representations, and the potential 
for quantum-like models to illuminate the inner workings of LLMs. Our central proposition is that the LLM 
embedding space can be viewed as a quantized semantic space, where individual words and phrases correspond to 
distinct quantum states. This perspective allows us to draw parallels between the probabilistic outputs of 
LLMs and the inherent uncertainties characteristic of quantum systems.

A key element of our approach is the extension of the standard, real-valued embedding space to the complex domain. 
This complexification enables us to model semantic interference effects, drawing direct parallels to phenomena such 
as the double-slit experiment in quantum mechanics. We demonstrate how this complex representation, combined with 
concepts like potential landscapes (e.g., the double-well potential for modeling semantic ambiguity) and path 
integral formalisms, can provide new insights into the behavior of LLMs, including their ability to generate 
creative text and handle nuanced semantic relationships.

By embracing this interdisciplinary approach, we aim to establish a theoretical foundation for 
understanding LLM behavior through the lens of quantum mechanics. We propose that LLM embedding 
spaces exhibit emergent properties analogous to quantum systems, 
potentially leading to new views in natural language understanding and generation.

\section{Background and Related Work}

LLMs have revolutionized Natural Language Processing (NLP), 
achieving state-of-the-art results in various tasks such as text generation, translation, 
and question answering. These models are typically based on the Transformer 
architecture \cite{transformer}, which utilizes self-attention mechanisms to 
capture long-range dependencies in text.  A key distinction of the Transformer architecture 
is its departure from previous recurrent neural network approaches.  Instead of recurrence, 
the Transformer relies entirely on attention mechanisms, enabling parallel processing of 
the input sequence and significantly improved training efficiency. The Transformer 
architecture has enabled LLMs to scale to unprecedented sizes, with models like 
BERT \citep{devlin} 
and GPT-3 \citep{brown} containing billions of parameters.

The self-attention mechanism is a core innovation of the Transformer. It allows 
the model to weigh the importance of different words in the input sequence when 
processing each word, effectively capturing contextual relationships. This is 
achieved by computing attention weights based on the relationships between 
"queries", "keys", and "values" derived from the input embeddings. The attention 
weights determine the contribution of each word to the representation of other words, 
allowing the model to focus on the most relevant parts of the input when making predictions. 
Furthermore, multi-head attention enhances this capability by allowing the model to attend to 
different aspects of the input in parallel, capturing a richer set of relationships.

A key component of LLMs is the embedding space, where words and phrases are represented 
as high-dimensional vectors. These embeddings are learned during the training process 
and capture semantic relationships between words. Words with similar meanings are 
located closer together in the embedding space, while dissimilar words are further apart. 
Understanding the structure and properties of these embedding spaces is crucial for 
interpreting the behavior of LLMs. The geometry of these embedding spaces has been explored 
by researchers like Mimno and Thompson \citep{mimno}. Furthermore, the 
way context influences these embeddings has been investigated by Tenney et al. \citep{tenney}.

Several approaches have been proposed for analyzing LLM embedding spaces. Dimensionality 
reduction techniques, such as Principal Component Analysis (PCA) \cite{PCA} and t-distributed 
Stochastic Neighbor Embedding (t-SNE) \cite{tSNE}, are often used to visualize the 
embedding space in lower dimensions. Geometric analysis techniques, such as calculating 
cosine similarity between vectors \cite{cosine}, are used to quantify the semantic similarity 
between words and phrases. These approaches have provided valuable insights into the organization 
of semantic information in LLMs. Mitchell and Lapata \citep{mitchell} have developed 
methods for evaluating the compositionality of these models, while Arora et al. \citep{arora} 
have explored the linear algebraic structure of word senses.

The connection between information theory and physics has been explored in various contexts. 
Shannon's information theory provides a framework for quantifying the amount of information 
contained in a message \cite{shannon}. Landauer's principle establishes a fundamental link between 
information and thermodynamics, stating that erasing one bit of information requires a minimum 
amount of energy dissipation \cite{landauer}. These concepts have been applied to the study of 
computation, complexity, and the limits of information processing.

The application of field theory to classical complex phenomena has also been successful. 
For example, field theory has been used to model turbulence \cite{laine}, providing a theoretical 
framework for understanding the statistical properties of turbulent flows. Furthermore, 
field-theoretic approaches are valuable in describing phase transitions in classical 
statistical mechanics \cite{cardy} and the emergent behavior of active 
matter systems. These applications demonstrate the power of field theory to describe 
emergent phenomena in complex systems.

Our work builds upon these existing approaches by exploring the analogy between LLM embedding 
spaces and quantum mechanics. Quantum mechanics, pioneered by Heisenberg \citep{heisenberg1925}, 
Schrödinger \citep{schrodinger}, and Dirac \citep{dirac1925}, provides a powerful framework 
for understanding the behavior of matter at the atomic and subatomic level \citep{merzbacher}. 
Dirac's work on 
quantum electrodynamics \citep{dirac1927, dirac1930} laid the foundation for quantum field theory 
\citep{ramond, kaku, bertkmann}, 
further developed by Schwinger \citep{schwinger}. The concept of symmetry \citep{coleman}, crucial 
in physics, was formalized by Wigner \citep{wigner} and plays a key role in understanding 
particle physics, including the Higgs mechanism \citep{higgs, englert, guralnik}. 
We propose a new approach based on quantum principles that leverages the theoretical frameworks of 
quantum mechanics to gain a deeper understanding of the organization and dynamics of semantic 
spaces. This approach is inspired by the observation that LLMs rely on a finite vocabulary of 
tokens, suggesting an underlying discreteness in the seemingly continuous semantic space. We 
extend the standard, real-valued embedding space to the complex domain to model semantic 
interference effects, drawing parallels to phenomena such as the double-slit experiment in 
quantum mechanics. Furthermore, we introduce a new, complex-valued similarity measure that 
captures both magnitude and phase information, offering a more nuanced comparison of semantic 
representations. We present a path integral formalism, based on a nonlinear Schrödinger 
equation with a gauge field and Mexican hat potential, for modeling the dynamics of LLM behavior. 
This builds on previous work exploring quantum-like models in cognition \citep{bruza, 
busemeyer, atmanspacher} and semantic composition \citep{widdows}.

By applying quantum mechanical concepts to the analysis of LLM embedding spaces, our approach aims 
to provide a fresh perspective on these complex systems. Building on previous work that has 
explored connections between information theory, physics, and natural language processing, 
this quantum-like framework has the potential to unlock new understandings of the emergent 
properties of LLMs and to inspire future research.

\section{Probabilistic LLM Behavior: The Coin Flipping Analogy}

To begin exploring the analogy between LLMs and quantum mechanics, we first acknowledge the 
inherent probabilistic nature of LLM outputs. While often perceived as deterministic systems trained 
to mimic human language, LLMs exhibit a degree of randomness, particularly evident in text generation. 
This stochasticity arises from the sampling process used to select the next token in a sequence, where 
probabilities are assigned to different tokens based on the model's learned distribution. To illustrate 
this probabilistic behavior and introduce relevant quantum mechanical concepts, we employ a simple coin-flipping analogy.
In the following example, and also in other examples in this article, we use the OpenAI embedding model and gpt-4o.

Consider the following prompt presented to an LLM: """You are a game machine. Choose a number between 0 and 1. 
If the number is less than 0.5, enter "heads". Otherwise, write "tails". Just answer "heads" or "tails".""" 
When presented with this prompt in a conversational setting (where the history is preserved), 
the LLM generates a sequence of responses that, over many iterations, approximates a fair coin flip. 
We observe roughly 50\% of the answers being "heads" and 50\% being "tails." This empirical observation  
suggests an underlying probabilistic mechanism governing the LLM's output, rather than a purely deterministic process.

In quantum mechanics, a two-level system provides a fundamental model for describing systems with two distinct 
and discrete states. Unlike classical systems, a quantum two-level system can exist in a superposition, a 
linear combination of both states. Familiar examples include the spin of a spin-1/2 particle (spin up or 
spin down) or the polarization of a photon (horizontal or vertical). The state of a general two-level 
system can be represented as a vector in a two-dimensional Hilbert space,

\begin{equation}
    \ket{\psi} = c_1\ket{1} + c_2\ket{2},
\end{equation}

\noindent
where $\ket{1}$ and $\ket{2}$ represent the two orthonormal basis states, and $c_1$ and $c_2$ are complex 
probability amplitudes. The probability of measuring the system in state $\ket{1}$ is given by $|c_1|^2$, 
and the probability of measuring the system in state $\ket{2}$ is given by $|c_2|^2$. These probabilities 
must sum to one, reflecting the certainty that the system is in one of the two possible states upon measurement,

\begin{equation}
    |c_1|^2 + |c_2|^2 = 1.
\end{equation}

\noindent
We can draw a conceptual analogy between the LLM coin-flipping example and a quantum two-level system. 
Let $\ket{\text{heads}}$ and $\ket{\text{tails}}$ represent the two possible states of the LLM's response, 
analogous to the spin-up and spin-down states of a spin-1/2 particle. The prompt can be viewed as 
an "interaction" or "measurement" that forces the LLM to "choose" between these two states. Prior to 
the prompt, the LLM can be considered to be in a superposition of these two states,

\begin{equation}
    \ket{\text{prompt}} = c_1\ket{\text{heads}} + c_2\ket{\text{tails}}.
\end{equation}

\noindent
In this analogy, $c_1$ and $c_2$ represent the probability amplitudes for the LLM to respond with "heads" or "tails", 
respectively. The observed frequencies of "heads" and "tails" in the LLM's output can then be interpreted as the 
probabilities of measuring the system in each state,

\begin{eqnarray}
    |\braket{\text{heads}}{\text{prompt}}|^2 &\approx& 0.5 \\
    |\braket{\text{tails}}{\text{prompt}}|^2 &\approx& 0.5
\end{eqnarray}

\noindent
While this analogy provides a useful and intuitive starting point for understanding the probabilistic 
nature of LLMs, it is crucial to recognize its inherent limitations. In the quantum mechanical example, 
$\ket{\text{heads}}$ and $\ket{\text{tails}}$ represent distinct and orthogonal quantum states, not 
simply vectors in a real-valued embedding space. Furthermore, the dynamics of a quantum system are 
governed by the Schrödinger equation, which describes the evolution of the wave function and the 
superposition of states. The act of measurement in quantum mechanics also leads to the collapse of 
the wave function into a single, definite state. These concepts are not directly mirrored in the 
standard LLM architecture. Furthermore, the standard LLM embedding space does not inherently 
support complex probability amplitudes, which are essential for describing quantum interference 
effects. These limitations motivate the need for a more sophisticated model that can capture the 
wave-like behavior of semantic representations and incorporate complex phases, as we will explore 
in the following sections.

\section{Core Quantum Mechanical Analogies}

It is crucial to first clarify the role of nonlinearity in the context of LLM embedding spaces. We distinguish between 
two key stages: (1) the LLM's training process and prediction calculation, which are inherently nonlinear due to the 
use of neural networks with nonlinear activation functions, and (2) the subsequent analysis of the learned embeddings, 
such as calculating cosine similarity between vectors. While the latter is a linear operation, it operates on vectors 
resulting from a nonlinear process, thus inheriting the effects of that nonlinearity.
Based on this distinction, we will explore two complementary approaches in this article: one that leverages the linearity 
of the embedding space when comparing results (embedding vectors, states) and another that explicitly incorporates 
nonlinearity during result calculation to capture more nuanced semantic phenomena.

To draw parallels with quantum mechanics, we adopt the following key assumptions:

\begin{enumerate}
    \item Completeness of Vocabulary: We assume that the LLM's finite vocabulary forms a (approximately) complete basis 
    for representing semantic information.  While this is not strictly true, we treat it as a reasonable approximation 
    for the purposes of our analogy.

    \item Semantic Space: Drawing an analogy to quantum mechanical phase space, we define a semantic space 
    as the complex extension of the embedding space (or, equivalently, the configuration space). 
    If the embedding space has dimension N, the semantic space has 2N real dimensions. This doubling 
    of dimensionality arises from the complex extension, which effectively provides 2 degrees of 
    freedom for each original dimension in the configuration space. This allows the modeling of 
    interference effects and other quantum-like phenomena within the semantic space. Analogous 
    to how time is treated in classical mechanics, the time taken for an LLM to generate a 
    prediction from a given input is considered separately from the semantic space itself. 
    While phase space describes the possible states of a system at a given time, and time 
    evolution traces a trajectory through phase space, the 'time' dimension for LLM prediction 
    generation is not included as a dimension within the semantic space. It is a parameter that 
    governs the evolution or trajectory through the semantic space.

    \item Quantized Semantic States: Semantic states are discrete and distinct, analogous to the quantization of energy 
    levels in quantum systems. This allows us to represent words and phrases as distinct quantum states within the 
    semantic space.
    The reliance of LLMs on a finite vocabulary of tokens provides a basis for this assumption, suggesting an underlying 
    discreteness in their semantic processing.

    \item Schrödinger Equation for Semantic Wave Propagation (Linear Model): In our first approach, we treat the semantic 
    space as linear. The evolution of the semantic wave function within this space is governed by the standard, 
    linear Schrödinger equation. This equation serves as a fundamental description of wave propagation for non-relativistic 
    particles in the absence of explicit nonlinear effects. This approximation is valid when considering basic 
    semantic relationships and allows us to leverage the superposition principle and plane wave solutions.

    \item Nonlinear Semantic Wave Propagation (Nonlinear Model): In our second, more sophisticated approach, we 
    explicitly account for the nonlinear nature of the embedding space. We model this nonlinearity through two 
    distinct mechanisms: (a) by introducing a cubic term directly into the Schrödinger equation, resulting in a 
    nonlinear Schrödinger equation (NLSE), and (b) by employing nonlinear potential functions. The cubic term allows 
    us to explore phenomena such as semantic self-interaction and the formation of semantic solitons. Nonlinear 
    potentials, such as the double-well potential, enable us to model semantic ambiguity and context-dependent 
    meaning. This approach is particularly relevant when considering complex semantic relationships and emergent phenomena.

    \item Semantic Charge and Interaction: We assume "semantic charge" as a property associated with 
    each word, phrase, or semantic concept within the LLM's vocabulary. (We will use this
    concept in later sections when discussing path integrals).
    This charge is represented 
    by the magnitude of the corresponding coefficient in the semantic wave function's expansion 
    in terms of basis states. The interaction between semantic charges is mediated by the gauge 
    field, analogous to how electromagnetic forces are mediated by photons. Specifically, regions 
    of high semantic charge density attract or repel each other based on the sign of the coupling 
    constant in the nonlinear Schrödinger equation or the form of the Mexican hat potential. 
    This interaction influences the overall semantic structure and coherence of the LLM's representation.

\end{enumerate}

\noindent
It is essential to emphasize that these 
are analogies, not literal physical equivalences.
They provide a theoretical framework for examining LLM behavior.
To manage the computational complexity introduced by nonlinearity in the nonlinear model, we employ approximations 
when dealing with wave function propagation. It's important to note that exact solutions to nonlinear equations 
are rare; therefore, approximations are frequently necessary in these contexts.

\section{Extending the Embedding Space: Semantic Interference and the Need for Complex Representations}

The standard LLM embedding space represents words, phrases, and sentences as real-valued vectors.
Cosine similarity, a real number between -1 and 1, then quantifies the semantic similarity between
these vectors based on the angle between them. While effective, this approach has limitations in capturing
more nuanced semantic relationships, particularly those involving contextual information and interference
effects. In this section, we propose a theoretical extension where the embedding space is complexified,
allowing us to represent "meaning" as a wave function, where the phase is crucial for capturing interference
effects analogous to those observed in quantum mechanics.

\subsection{Limitations of Real-Valued Embeddings: The Case for Contextual Sensitivity} 

To illustrate the limitations of the real-valued embedding space and motivate the need for complex
representations, consider the task of categorizing LLM entries or distinguishing nuanced differences 
in word usage.
Suppose we want to determine whether a given sentence is related to "dogs" or "cats," or, more subtly,
to differentiate between words with similar overall meaning but different contextual appropriateness.
We might have the following prompts:

\begin{eqnarray}
    \text{prompt1} &=& \text{"I want to talk about cats."} \\
    \text{prompt2} &=& \text{"I like cats."} \\
    \text{prompt3} &=& \text{"I am afraid of dogs."} \\
    \text{prompt4} &=& \text{"Dogs are more loyal than cats."}
\end{eqnarray}

\noindent
The standard approach is to calculate the cosine similarity between the embedding of each prompt and the
embeddings of "dogs" and "cats." 

\begin{figure}[h!]
    \centering
    \includegraphics[width=0.8\textwidth]{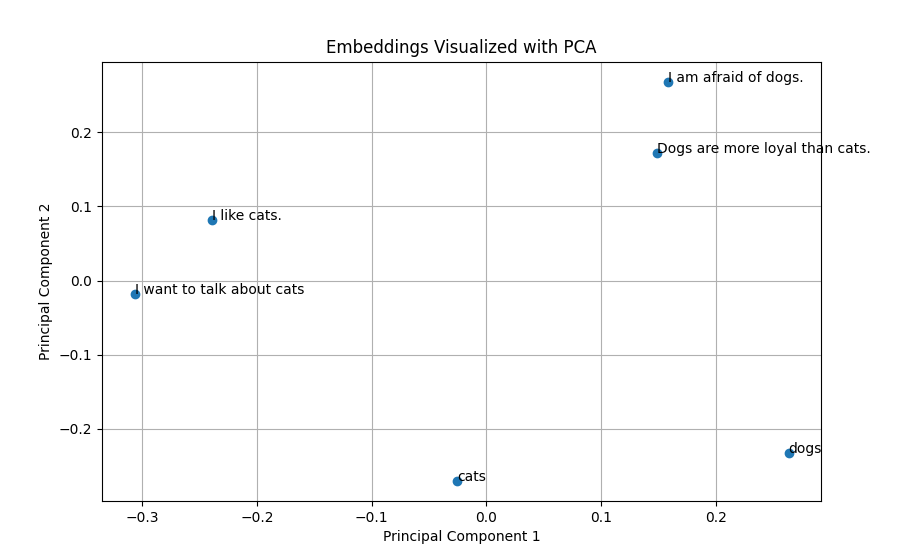}
    \caption{Principal Component Analysis (PCA) projection of word embeddings for "dogs," "cats," and example prompts. 
    This visualization demonstrates the limitations of real-valued embeddings and cosine similarity in capturing nuanced 
    semantic relationships. While prompts related to "cats" generally cluster closer to the "cats" embedding, the overlap 
    between clusters suggests that context and subtle differences in meaning are not fully captured by this approach.}
    \label{fig:PCA}
\end{figure}

Given two vectors $\vec{A}$ and $\vec{B}$, their inner product is defined as

\begin{equation}
   \vec{A}\cdot \vec{B} =  |\vec{A}||\vec{B}|\cos\theta,
\end{equation}

\noindent
from which we obtain the definition of cosine similarity,

\begin{equation}
    S_C(\vec{A},\vec{B})= \cos\theta = \frac{\vec{A}\cdot \vec{B}}{|\vec{A}||\vec{B}|}.
\end{equation}

\noindent
The values are normalized are therefore between -1 to 1. Value 1 corresponds exactly aligned (similar) vector
and -1 the most non-similar.
In vector embedding space cosine similarity gives a useful measure of how similar two words, sentence or 
documents are likely to be,
in terms of their subject matter, and independently of the length of the documents.
Cosine similarity provides a useful measure of semantic similarity, but it only captures the angle between the vectors.
It does not account for potential interference effects that might arise from the superposition of different
semantic components, nor does it fully capture nuanced differences in word usage.

In this example, we might obtain the following cosine similarities:

\begin{eqnarray}
    &&S_C(\text{prompt1},\text{"dogs"}) =  0.7924637278190193 \\
    &&S_C(\text{prompt1},\text{"cats"}) =  0.8691721116453197 \\
    &&S_C(\text{prompt2},\text{"dogs"}) =  0.7948651093796374 \\
    &&S_C(\text{prompt2},\text{"cats"}) =  0.8742215747399027 \\
    &&S_C(\text{prompt3},\text{"dogs"}) =  0.8292989582958668 \\
    &&S_C(\text{prompt3},\text{"cats"}) =  0.7966300696893088\\
    &&S_C(\text{prompt4},\text{"dogs"}) =  0.8295814659291845 \\
    &&S_C(\text{prompt4},\text{"cats"}) =  0.8215583147667265
\end{eqnarray}

\noindent
We may visualize these embeddings using principal component analysis (PCA), see Figure \ref{fig:PCA}. Semantically
similar words and sentencens are closer to each other. 

\subsection{Semantic Interference: Drawing Parallels to the Double-Slit Experiment} 

\begin{figure}[h!]
    \centering
    \includegraphics[width=0.8\textwidth]{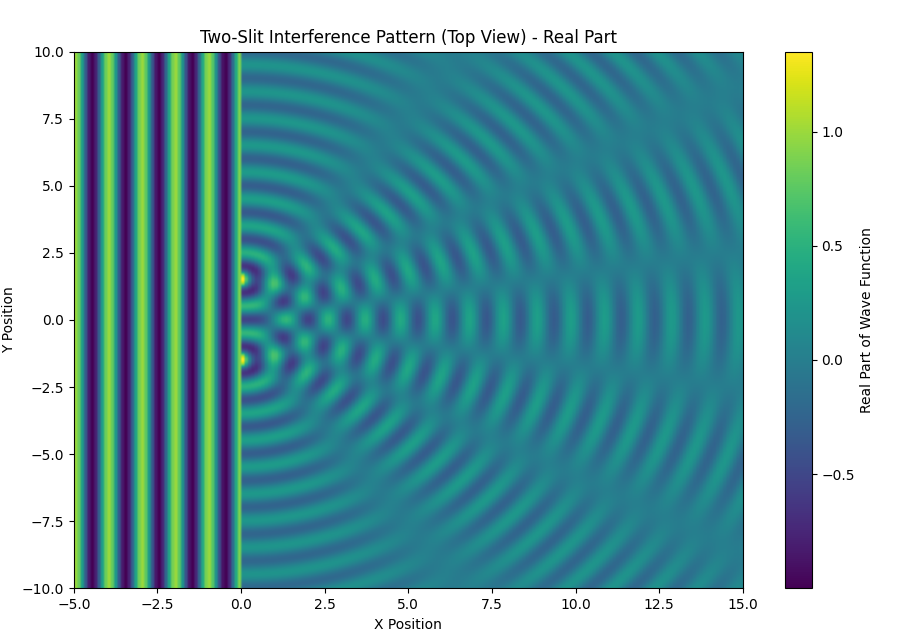}
    \caption{The double-slit experiment, illustrating the importance of phase information in capturing interference effects. 
    In analogy to LLMs, this demonstrates why complex-valued representations are necessary to model the 
    nuanced semantic relationships that cannot be captured by real-valued embeddings alone.}
    \label{fig:2-slit}
\end{figure}

To capture potential interference effects within the semantic space of LLMs, we draw 
an analogy to the double-slit experiment in quantum mechanics, see Figure \ref{fig:2-slit}. 

In the standard double-slit experiment, particles (e.g., photons or electrons) are directed towards a screen with two slits. 
The resulting pattern on a detector screen behind the slits exhibits an interference pattern, demonstrating the 
wave-like nature of the particles. The probability distribution on the screen is not simply the sum of the 
probabilities from each slit individually; instead, it displays interference fringes due to the superposition of 
the waves emanating from each slit.

The wave functions of the particles passing through each slit can be written as

\begin{eqnarray}
    \psi_1(x,y) &=& A_1(x,y) \exp(i\varphi_1), \\
    \psi_2(x,y) &=& A_2(x,y) \exp(i\varphi_2),
\end{eqnarray}

\noindent
where $A_1(x,y)$ and $A_2(x,y)$ represent the amplitudes, and $\varphi_1$ and $\varphi_2$ represent the phases 
of the waves at a point $(x, y)$ on the detector screen. The probability distribution is then given by

\begin{equation}
    P(x,y) = |\psi_1(x,y) + \psi_2(x,y)|^2 = |A_1(x,y)|^2 + |A_2(x,y)|^2 
    + 2 |A_1(x,y)| |A_2(x,y)| \cos(\varphi_1 - \varphi_2).
\end{equation}

\noindent
The crucial term $2 |A_1(x,y)| |A_2(x,y)| \cos(\varphi_1 - \varphi_2)$ is the "interference term," which arises 
from the superposition of the complex probability amplitudes. A purely real formulation lacks the phase 
information necessary to describe this interference. The double-slit experiment demonstrates the 
wave-particle duality of light and matter. Both wave and particle descriptions are valid, 
and the theoretical framework used to describe the phenomenon depends on the experimental setup.

\subsection{Extending the Analogy: N-Dimensional Semantic Space} 

We extend our framework to a higher-dimensional space. Consider an analogous experiment conducted within 
a phase space of $2N$ real dimensions, where $N$ represents the dimensionality of the LLM embedding space. An additional 
dimension represents time. This higher-dimensional space is designed to capture the intricate relationships between 
semantic concepts. In this analogy, each "slit" corresponds to a distinct semantic context or a particular facet 
of a word's meaning, and the wave function or "particle" that traverses the slit represents a word, sentence, or 
paragraph under investigation.

Given that we are currently focusing on the utilization of pre-trained embeddings, rather than the training process 
itself, we are operating within a regime where the explicit nonlinearities of the training process are not directly 
considered. Therefore, we treat the semantic space as approximately linear in this context. Consequently, the total 
field at any point can be approximated as the linear superposition of the individual fields originating from each 
"slit." This allows us to model the semantic wave propagation using plane waves and the principle of superposition.

Under these assumptions, we consider a simplified two-slit scenario where two wave functions, denoted as $\psi_1$ 
and $\psi_2$, traverse the slits. The total wave function is then given by the superposition

\begin{equation}
    \psi = \psi_1 + \psi_2.
\end{equation}

\noindent
Approximating these waves as plane waves, we express them as

\begin{align}
    \psi_1(t, \mathbf{x}) &= A_1 e^{i(\mathbf{k}_1 \cdot \mathbf{x} - \omega_1 t + \varphi_1 )}, \\
    \psi_2(t, \mathbf{x}) &= A_2 e^{i(\mathbf{k}_2 \cdot \mathbf{x} - \omega_2 t + \varphi_2 )},
\end{align}

\noindent
where $A_1$ and $A_2$ represent the amplitudes, $\mathbf{k}_1$ and $\mathbf{k}_2$ are the wave vectors
in high-dimensional space, 
$\omega_1$ and $\omega_2$ are the angular frequencies, $\varphi_1$ and $\varphi_2$ are the initial phases.
The intensity of the wave function, which can be interpreted as the probability density, is then

\begin{equation}
    P = |\psi|^2 = |\psi_1 + \psi_2|^2 = |\psi_1|^2 + |\psi_2|^2 + 2 |\psi_1| |\psi_2| \cos(\varphi_1 - \varphi_2 
     + \mathbf{k}_1 \cdot \mathbf{x} 
    - \mathbf{k}_2 \cdot \mathbf{x} - (\omega_1 - \omega_2)t).
\end{equation}

\noindent
For simplicity, we assume that $\varphi_1 = 0$  and 
consider the case where $\omega_1 = \omega_2 = \omega$. This simplifies the expression to

\begin{equation}
    P = |A_1|^2 + |A_2|^2 + 2 |A_1| |A_2| \cos(-\varphi_2 + (\mathbf{k}_1 - \mathbf{k}_2) \cdot \mathbf{x}).
\end{equation}

\noindent
Our objective is to relate the term $\cos(-\varphi_2 + (\mathbf{k}_1 - \mathbf{k}_2) \cdot \mathbf{x})$ to 
the cosine similarity between the embedding vectors $\mathbf{v}_1$ and $\mathbf{v}_2$.
We assume that the wave vectors are proportional to the embedding vectors,

\begin{equation}
    \mathbf{k}_1 = \alpha \mathbf{v}_1, \quad \mathbf{k}_2 = \alpha \mathbf{v}_2,
\end{equation}

\noindent
where $\alpha$ is a proportionality constant.
Then

\begin{equation}
    P = |A_1|^2 + |A_2|^2 + 2 |A_1| |A_2| \cos(-\varphi_2 + \alpha (\mathbf{v}_1 - \mathbf{v}_2) \cdot \mathbf{x}).
\end{equation}

\noindent
Now, assuming that we are evaluating the intensity at the origin, $\mathbf{x} = \mathbf{0}$, we have

\begin{equation}
    P = |A_1|^2 + |A_2|^2 + 2 |A_1| |A_2| \cos(-\varphi_2).
\end{equation}

\noindent
To relate $\varphi_2$ to the cosine similarity, we assume that the initial phase $\varphi_2$ is related to the 
angle between the embedding vectors. Specifically, $\varphi_2 = \beta \theta$,
where $\theta$ is the angle between $\mathbf{v}_1$ and $\mathbf{v}_2$, and $\beta$ is another 
proportionality constant. Then

\begin{equation}
    P = |A_1|^2 + |A_2|^2 + 2 |A_1| |A_2| \cos(-\beta \theta).
\end{equation}

\noindent
Given that cosine similarity is defined as $\text{cos}(\theta)
 = (\mathbf{v}_1 \cdot \mathbf{v}_2)/(||\mathbf{v}_1||||\mathbf{v}_2||)$, we can write

\begin{equation}
    P = |A_1|^2 + |A_2|^2 + 2 |A_1| |A_2| \cos\left(-\beta \arccos\left(\frac{\mathbf{v}_1 
    \cdot \mathbf{v}_2}{||\mathbf{v}_1||||\mathbf{v}_2||}\right)\right).
\end{equation}

\noindent
This simplified formula provides an approximate means of connecting the intensity (probability of a semantic 
interpretation) 
to the cosine similarity between embedding vectors, while also incorporating phase information. The key point is that 
the phase difference between the semantic waves plays a crucial role in determining the overall intensity, 
analogous to its role in the double-slit experiment. This suggests that a complex-valued representation, 
which includes phase, is essential for capturing the nuances of semantic meaning.

The interpretation is similar to the 2D case, but now the interference pattern exists in a much higher-dimensional space. 
The phase difference $\varphi_1(\mathbf{x})-\varphi_2(\mathbf{x})$ captures the relative semantic relationships between 
the two "slits" (semantic contexts) at a particular point in the N -dimensional space. 

\subsection{Semantic States as Superpositions: Towards Complex Representations} 

In the context of LLMs, we can think of different semantic components of a prompt as analogous to the waves emanating 
from the slits. The overall meaning of the prompt is then determined by the superposition of these semantic 
components, which can lead to constructive or destructive interference. This motivates us to extend the 
real-valued LLM embedding space to the complex domain. In this complexified space, each word, phrase, or 
sentence is represented by a complex vector, and the semantic relationships between them are determined 
by both the magnitudes and the phases of these vectors.

For example, we could represent "prompt1" as

\begin{equation}
    \ket{\text{prompt1}} =
    |c_1|\exp(i\varphi_1)\ket{\text{dogs}} +
    |c_2|\exp(i\varphi_2)\ket{\text{cats}} + \sum_{i\ne 1,2}  |c_i|\exp(i\varphi_i)\ket{\psi_i},
\end{equation}

\noindent
where $|c_i|$ represents the magnitude of the contribution of each basis state $\ket{\psi_i}$ 
(e.g., "dogs," "cats," or other semantic components), and $\varphi_i$ represents the phase. 
The phase information is crucial for capturing the relative relationships between different 
semantic components; for instance, synonyms might have similar magnitudes and similar phases, 
while antonyms might have similar magnitudes but opposite phases. 

\subsection{Illustrative Example: Semantic Clustering and Similarity} 

To further illustrate the potential of complex embeddings, consider a simplified program that scans 
the alphabet to find words with close similarity scores for words "dog" and "cat". The script gives
the cosine similarity difference as

\begin{equation}
  \Delta d = |S_C(\text{'Letter'}, \text{dog})-S_C(\text{'Letter'},\text{cat})|.
\end{equation}

\noindent
As an example (Letter:, "alphabet", $\Delta d$):

\begin{verbatim}
Letter: a  0.03202649985101025
Letter: b  0.023594058171430343
Letter: c  0.017036352475257588
Letter: t  0.012467653389398925
Best letter for length 1: t with difference 0.012467653389398925
Letter: aa  0.021800746027967888
Letter: ab  0.003948614333593059
Letter: aq  0.002333507704697757
Letter: bk  0.0016448531528867605
Letter: cp  0.0009932341767737718
Letter: kh  0.0002932433299662751
Letter: mc  0.0002022452982680667
Best letter for length 2: mc with difference 0.0002022452982680667
Letter: aaa  0.016703392191957933
Letter: aab  0.014940121087262392
Letter: aac  0.00106761509332709
Letter: acx  6.381206526662186e-05
Letter: cas  1.5506346889848643e-06
Best letter for length 3: cas with difference 1.5506346889848643e-06
\end{verbatim}

\noindent
We observe that as the number of letters increases, groups of three words tend to exhibit tighter semantic clustering, 
with two of the words being semantically closer to the third. This increased clustering is reflected in a lower cosine 
similarity between the word pairs (indicating greater similarity), a trend that presumably continues with further 
increases in letter count. It's also possible that longer words lead to the formation of a greater number of 
distinct semantic 
clusters, which are themselves relatively distant from each other in the overall semantic space.

The program identifies "cas" as being very similar to "dog" and "cat" according to some real-valued similarity metric.  
We might represent this in our complexified space as

\begin{equation}
    \ket{\text{cas}} =   |c_1|\exp(i\varphi_1)\ket{\text{dog}}   +
    |c_2|\exp(i\varphi_2)\ket{\text{cat}} + \sum_{i \ne 1,2} |c_i|\exp(i\varphi_i) \ket{\psi_i},
\end{equation}

\noindent
where, for the sake of example, we have the same cosine similarity for both phases,

\begin{equation}
    \varphi_1 = 0.82422 \quad \varphi_2 = 0.82422
\end{equation}

\noindent
This output suggests that 'cas' and 'acx' exhibit a high degree of similarity within the 
embedding space. However, cosine similarity alone fails to capture the specific differences 
in meaning and usage that differentiate these terms. While both might relate to the general 
concept of a 'situation,' their specific connotations and contextual appropriateness may vary 
significantly.

In a complexified embedding space, the phase component could encode these subtle distinctions. For instance, 
if we consider the word 'situation' as a superposition of basis states, the phase relationships between 
'situation' and 'cas' might differ from those between 'situation' and 'acx,' reflecting their varying 
degrees of formality or technicality. This phase information, absent in real-valued embeddings, could 
provide a more refined measure of semantic similarity, enabling LLMs to generate more contextually 
appropriate and nuanced text.

\subsection{The Mexican Hat Potential: A Geometrical Interpretation of Semantic Similarity} 

The cosine similarity in the real-valued embedding space can be seen as a projection of the complex 
similarity onto the real axis. The phase difference between two complex vectors then determines the 
value of the cosine similarity.

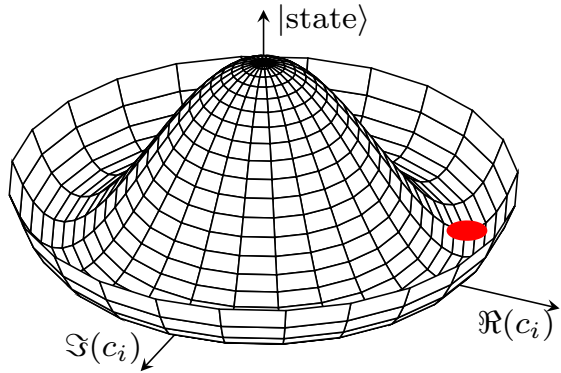
\begin{figure}[h!]
    \centering
    \begin{tikzpicture}[font=\footnotesize, scale=1.5]
        \begin{axis}[
            axis lines=center,
            axis equal,
            domain=0:360,
            y domain=0:1.25,
            y axis line style=stealth-,
            y label style={at={(0.35,0.18)}},
            xmax=1.6,zmax=1.3,
            xlabel = $\Re(c_i)$,
            xlabel style={yshift=-0.6cm},
            ylabel=$\Im{(c_i)}$,
            zlabel=$\ket{\text{state}}$,
            ticks=none
        ]
        \addplot3 [surf,shader=flat,draw=black,fill=white,z buffer=sort] ({sin(x)*y}, {cos(x)*y}, {(y^2-1)^2});
        \coordinate (center) at (axis cs:0,0,1);
        \coordinate (minimum) at (axis cs:{cos(30)},{sin(30)},0);
        \fill[red] (minimum) circle (0.1);
        \end{axis}
    \end{tikzpicture}
    \caption{The Mexican hat potential, illustrating the concept of spontaneous symmetry breaking. In the context of 
    LLMs, this potential can be used to model the emergence of stable semantic meanings, where the system "chooses" 
    a particular interpretation despite the potential's inherent symmetry.}
    \label{fig:mexican-hat} 
\end{figure}

To preserve the continuous nature of the similarity measure and ensure that it remains bounded between -1 and 1, we 
could also consider a potential with circular symmetry in the complex plane. A potential that exhibits this symmetry 
is the Mexican hat potential, which allows for a continuous range of phase values while maintaining a 
well-defined minimum energy state, see Figure \ref{fig:mexican-hat}. The magnitude $|c_i|$ of the coefficient 
represents the "distance" 
from the center of the hat. A larger magnitude means the state is further away from the center. 
Parameter $\varphi_i$ can have values $[0,2\pi]$ and it represents the angle around the circle at the bottom of the hat.

The Mexican hat potential is crucial in quantum mechanics and quantum field theory for illustrating spontaneous 
symmetry breaking, where the system selects a specific vacuum state despite the potential's symmetry. This symmetry 
breaking leads to Goldstone bosons or, via the Higgs mechanism, massive gauge bosons, which are essential for 
understanding fundamental phenomena in particle physics, superconductivity, and cosmology.

\section{Semantic Wave Function: A Quantum Mechanical Formulation of Meaning}

In this section, we introduce a quantum mechanical formulation of meaning within LLMs, 
building upon the analogy of a quantized semantic space. Our goal is to develop a tractable and 
interpretable model for understanding certain aspects of LLM behavior, such as the superposition of 
semantic states and the potential for interference effects. While we acknowledge that LLMs are 
inherently nonlinear systems, we choose to focus on linear calculations in this section in 
order to simplify the mathematical analysis and to gain a more intuitive understanding of 
these key phenomena. We recognize that this simplification limits the scope of the model and 
that it may not be able to capture all aspects of LLM behavior. However, we believe that this 
linear formulation provides a valuable starting point for exploring the potential connections 
between quantum mechanics and natural language processing, and that it can serve as a foundation 
for future extensions that incorporate nonlinearity.

The foundation of our analogy lies in the observation that LLMs operate with a finite vocabulary. This means
that the embedding space is not a continuous space, but rather a discrete space built upon a finite set of
words or tokens. We propose that each word or token in the LLM's vocabulary corresponds to a basis state in
a quantum system. We denote these basis states as $\ket{\psi_i}$, where $\psi_i$ represents the i-th word or
token in the vocabulary. These basis states form a discrete and (ideally) complete basis for representing semantic
information within the LLM. While the vocabulary is finite and therefore not truly complete in a mathematical
sense, we can treat it as an approximation of a complete basis for the purposes of our analogy. The embedding
vectors learned by the LLM can be seen as a representation of these basis states in a particular coordinate
system. In other words, the embedding vector for a word $\psi_i$ is a vector that specifies the "direction"
of the basis state $\ket{\psi_i}$ in the high-dimensional embedding space.

To formalize this, we define a semantic linear operator $\tilde{\Psi}(W, C)$, where $W$ represents a word or phrase,
and $C$ represents the context in which it is used. We postulate that this operator acts on states
$\ket{\psi_i}$ in the following way,

\begin{equation}
    \tilde{\Psi}(W, C)\ket{\psi_i}  = A_i(W, C)\ket{\psi_i}.
\end{equation}

\noindent
Here, $A_i(W, C)$ is the eigenvalue corresponding to the eigenstate $\ket{\psi_i}$. This eigenvalue represents
the amplitude for the word $W$ to be associated with the semantic component represented by $\ket{\psi_i}$ in
the context $C$. The eigenstates $\ket{\psi_i}$ form a complete basis for the semantic space, meaning that
any semantic state can be expressed as a linear combination of these eigenstates,

\begin{equation}
    \ket{\alpha} = \sum_{i}  c_i\ket{\psi_i},
\end{equation}

\noindent
where $c_i$ are complex coefficients that determine the contribution of each basis state to the overall
meaning. The magnitude of $c_i$ represents the strength of that aspect of the word's
meaning in the given context, and the phase of $c_i$ encodes additional semantic information, as discussed
in the previous section.

We can also express the semantic state in terms of the eigenstates of the semantic operator,

\begin{equation}
\ket{S(W, C)} = \tilde{\Psi}(W, C) \ket{\alpha}  = \sum_{i}  c_i A_i(W, C) \ket{\psi_i},
\end{equation}

\noindent
where $\ket{\alpha}$ is an arbitrary initial state. This equation shows how the semantic operator transforms
the initial state into the semantic state $\ket{S(W, C)}$ by weighting each eigenstate with its corresponding
eigenvalue and coefficient.

It is important to note that the coefficients $c_i$ are complex numbers, reflecting the extension of the
embedding space to the complex domain, as discussed in the previous section. While the standard LLM embedding
space is real-valued, we argue that allowing $c_i$ to be complex enables us to capture more nuanced semantic
relationships and interference effects. If we are measuring the state probabilities, they are in any case
always treated as absolute values like $| c_i|$.

\subsection{Example: Probabilistic Response Generation}

To illustrate this formulation, consider the following example. The system prompt is "You are a helpful AI assistant."
The user prompt is """Sun is shining. It is morning. It is dark. Is the sky blue? Answer "yes" or "no".
Use in output only words "yes" or "no" """. We make two tests. First with the temperature = 0.9 and top p = 1.0.
The expected state is

\begin{equation}
    \ket{\text{prompt}} =
    c_1\ket{\text{yes}} +
    c_2\ket{\text{no}} + \sum_{i\ne 1,2}  c_i A_i(W,C) \ket{\psi_i}.
\end{equation}

\noindent
We enter the prompt 10 times. The result is:

\begin{eqnarray}
    &&|\braket{\text{No}}{\text{prompt}}|^2 =  0.3\\
    &&|\braket{\text{No.}}{\text{prompt}}|^2 = 0.6 \\
    &&|\braket{\text{no}}{\text{prompt}}|^2 = 0.1
\end{eqnarray}

\noindent
In the next test, the temperature = 0.1 and top p = 1.0. The result is

\begin{eqnarray}
    &&|\braket{\text{No.}}{\text{prompt}}|^2 = 0.9 \\
    &&|\braket{\text{no}}{\text{prompt}}|^2 = 0.1
\end{eqnarray}

\noindent
This example demonstrates how the LLM's response can be interpreted as a superposition of basis states,
with the coefficients reflecting the probabilities of each state.

\subsection{Semantic Distinctness: Cosine Similarity Analysis}

To further illustrate the distinctness of semantic states, we perform a cosine similarity 
analysis on a set of related words. Using the OpenAI embeddings model, we calculate the cosine 
similarity between the word "no" and several other words, including variations in capitalization 
and punctuation, as well as semantically related and unrelated terms. The results are as follows:

\begin{eqnarray}
    &&S_C(\text{"no"},\text{"no"}) =  0.9999999999999998 \\
    &&S_C(\text{"no"},\text{"No"}) =  0.9694166470120369 \\
    &&S_C(\text{"no"},\text{"no."}) =  0.9020864046438366 \\
    &&S_C(\text{"no"},\text{"nobody"}) =  0.8278024059292596 \\
    &&S_C(\text{"no"},\text{"mono"}) =  0.8001234585682817 
\end{eqnarray}

\noindent
As the results demonstrate, only $S_C(\text{"no"},\text{"no"})$ achieves a cosine similarity of 1, 
indicating perfect alignment in the embedding space. All other words exhibit lower similarity scores, 
highlighting their distinct semantic states. Even minor variations in capitalization or punctuation 
result in deviations from perfect similarity. This analysis underscores the concept that each word, 
even those closely related in meaning or form, occupies a unique position in the semantic space and 
represents a distinct basis state.

\section{Modeling Semantic Ambiguity: The Double-Well Potential}

In the previous section, we introduced the concept of a semantic wave function to represent the meaning of a 
word or phrase in a given context. However, words often have multiple meanings, and the appropriate meaning 
depends on the context in which the word is used. To model this semantic ambiguity, we turn to the concept 
of a double-well potential from quantum mechanics.

\begin{figure}[h!]
    \centering
    \includegraphics[width=0.8\textwidth]{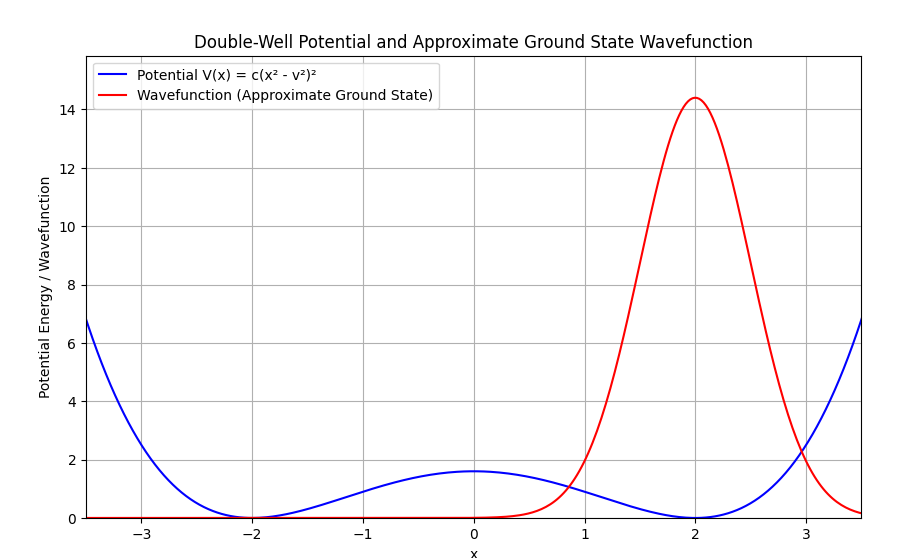}
    \caption{The double-well potential, illustrating how context can influence the interpretation of a 
    word in LLMs. The two minima represent different meanings, and the "tunneling" effect represents the 
    LLM's ability to switch between these meanings based on the surrounding context.}
    \label{fig:dw-potential}
\end{figure}

A double-well potential is a potential energy function with two minima separated by a barrier, see 
Figure \ref{fig:dw-potential}. 
While not directly analogous to the Mexican hat potential discussed earlier, it provides a useful 
framework for understanding systems with two stable states and transitions between them. Double-well 
potentials appear in diverse fields like molecular physics and condensed matter, and their importance 
lies in illustrating fundamental concepts of symmetry breaking and quantum tunneling in a simplified, tractable system.

A typical mathematical form for a double-well potential is

\begin{equation}
    V(x) = c(x^2-v^2)^2,
\end{equation}

\noindent
where $c$ and $v$ are positive constants. This potential has minima at $x = \pm v$.

The key characteristic of the double-well potential is that a particle placed in this potential will tend to settle into
either minimum. Even though the potential is symmetric under reflection ($x \rightarrow -x$), the particle "chooses" 
one of the two minima, effectively breaking the symmetry. Furthermore, in quantum mechanics, a particle can tunnel through 
the potential barrier separating the two minima. This means that the particle can transition from one minimum to the other, 
even if it doesn't have enough energy to overcome the barrier classically.

We propose that the double-well potential can be used to model semantic ambiguity in LLMs. Consider a word like "bank," 
which can refer to a financial institution or the side of a river. We can represent these two meanings as the two minima 
of a double-well potential. The context in which the word "bank" is used then determines which minimum the semantic wave 
function will settle into.

For example, in the sentence "The bank is open today," the context suggests that "bank" refers to a financial institution. 
The semantic wave function for "bank" in this context, which we can denote as $\ket{S(\text{"bank"}, C_1)}$, 
will have a high amplitude in the minimum corresponding to the financial institution meaning. 
Conversely, in the sentence "The bank of the river is eroding," the context suggests that "bank" 
refers to the side of a river. The semantic wave function for "bank" in this context, which we can denote 
as $\ket{S(\text{"bank"}, C_2)}$, will have a high amplitude in the minimum corresponding to the riverbank meaning.

The ability of a particle to tunnel between the two minima in a double-well potential can be seen as analogous to the 
LLM's ability to switch between different interpretations of a word or phrase depending on the context. A slight change 
in context can be seen as a perturbation that alters the potential landscape, causing the semantic wave function to tunnel 
from one minimum to the other.

To illustrate this, we can consider how changes in context $C$ affect the distribution of meanings. 
This is similar to how external potentials or interactions can alter the wave function in quantum mechanics. 
Suppose the context $C$ changes slightly, such as by adding a new word or altering the sentence structure. This change 
can be represented as a perturbation $\Delta c$, leading to a change in the semantic wave function,

\begin{equation}
\ket{S'(W, C + \Delta c)} = \sum_{i} \left(c_i + \delta c_i\right)\ket{\psi_i},
\end{equation}

\noindent
where $\delta c_i$ represents the change in the coefficient of the basis state $\ket{\psi_i}$ due to the context 
variation $\Delta c$.

In summary, the double-well potential provides a useful metaphor for understanding how LLMs handle semantic ambiguity. 
By representing different meanings of a word as different minima in a potential landscape, we can leverage the mathematical 
tools of quantum mechanics to analyze and understand the context-dependent nature of word meanings. The ability of the 
semantic wave function to tunnel between these minima reflects the LLM's ability to switch between different interpretations 
based on the surrounding context.

\section{A Complex-Valued Similarity Measure for Semantic Representations}

In previous sections, we established the framework for representing semantic information using complex-valued wave 
functions. This representation allows us to capture not only the magnitude but also the phase of each semantic component, 
enabling the modeling of interference effects and more nuanced semantic relationships. In this section, we 
introduce a new similarity measure that takes into account both the magnitudes and the phases of these 
complex-valued semantic representations.

Traditional cosine similarity, which operates on real-valued vectors, only captures the angular relationship 
between two vectors. It disregards the phase information that is crucial for capturing semantic interference. 
To address this limitation, we define a complex-valued similarity measure that incorporates both the magnitudes 
and the phase differences between the complex coefficients of the semantic wave functions.

Let $\ket{\text{text1}}$ and $\ket{\text{text2}}$ be two texts represented as superpositions of semantic concepts,

\begin{eqnarray}
    \ket{\text{text1}} &=& \sum_{i} c_{1i} \ket{\psi_i} = \sum_{i} |c_{1i}| e^{i\varphi_{1i}} \ket{\psi_i}, \\
    \ket{\text{text2}} &=& \sum_{i} c_{2i} \ket{\psi_i} = \sum_{i} |c_{2i}| e^{i\varphi_{2i}} \ket{\psi_i},
\end{eqnarray}

\noindent
where $c_{1i}$ and $c_{2i}$ are the complex coefficients, $|c_{1i}|$ and $|c_{2i}|$ are their magnitudes, 
and $\varphi_{1i}$ and $\varphi_{2i}$ are their phases. The basis states $\ket{\psi_i}$ represent the 
underlying semantic concepts.

We define the complex-valued similarity measure as

\begin{equation}
  S_T (\text{text1},\text{text2}) =  \sum_{i} c_{1i}^* c_{2i} 
  = \sum_{i} |c_{1i}| |c_{2i}| e^{i(\varphi_{2i}-\varphi_{1i})},
\end{equation}

\noindent
where $c_{1i}^*$ is the complex conjugate of $c_{1i}$. This similarity measure sums the products of the 
complex coefficients for each basis state. The magnitude of the resulting complex number provides a 
measure of the overall similarity between the two texts, while the phase provides information about 
their relative orientation in the semantic space.

To illustrate this, consider a simplified example with two semantic concepts: "happy" and "sad." Let

\begin{eqnarray}
  &&\ket{\psi_1} = \ket{\text{happy}}\\
  &&\ket{\psi_2} = \ket{\text{sad}}
\end{eqnarray}

\noindent
and consider the following texts

\begin{eqnarray}
    &&\ket{\text{text1}} = \text{"I am happy"}\\
    &&\ket{\text{text2}} = \text{"I am not happy"}
\end{eqnarray}

\noindent
We can represent these texts as (with example values)

\begin{eqnarray}
    &&\ket{\text{text1}} = 0.9 e^{i0.1}\ket{\text{happy}} + 0.1 e^{i3.0}\ket{\text{sad}}\\
    &&\ket{\text{text2}} = 0.1 e^{i0.3}\ket{\text{happy}} + 0.9 e^{i3.1}\ket{\text{sad}}
\end{eqnarray}

\noindent
The phase values are in radians.
Using our complex-valued similarity measure, we obtain

\begin{eqnarray}
  S_T(\text{text1},\text{text2}) &=& (0.9 e^{-i0.1})(0.1 e^{i0.3}) + (0.1 e^{-i3.0})(0.9 e^{i3.1}) \\
  |S_T(\text{text1},\text{text2}) | &\approx& 0.1798
\end{eqnarray}

\noindent
This value is relatively low, indicating that the two texts are not very similar according to this measure. 
This makes sense because one text expresses happiness, while the other expresses a lack of happiness. 
The phase of the similarity provides information about the relative orientation of the two texts in the semantic space.

This example demonstrates how the complex-valued similarity measure can capture subtle semantic differences 
that are not captured by traditional cosine similarity. By taking into account both the magnitudes and the 
phases of the complex coefficients, this measure provides a more nuanced and informative way to compare semantic 
representations.

\section{Semantic Dynamics: A Path Integral Approach}

In previous sections, we established a framework for representing semantic information using complex-valued 
wave functions and explored the concept of potential landscapes to model semantic ambiguity. To further 
understand the dynamics of semantic representations in LLMs, we now introduce a path integral formulation. 
This formalism allows us to model the evolution of the semantic wave function over time, taking into 
account the various factors that influence its behavior.

The path integral formulation is based on the idea that the probability amplitude for a system to evolve 
from an initial state to a final state is given by the sum over all possible paths connecting those two states. 
The generating functional is 

\begin{equation}
    Z = \int  \mathcal{D}[\psi] \mathcal{D}[\psi^*]   
     \exp\left(i S[\psi, \psi^*] \right),
\end{equation}

\noindent 
where the action is defined as

\begin{equation}
    S[\psi, \psi^*] = \int dt \int d^N x \mathcal{L}[\psi, \psi^*].
\end{equation}

\noindent
The Langrangian density $\mathcal{L}[\psi, \psi^*]$ defines the dynamics of a system
in $N$-dimensional phase.
In our context, the "system" is the semantic wave function, the "initial state" is the initial meaning of a prompt, 
and the "final state" is the LLM's response. Our goal is to derive an "effective" action that describes the 
dynamics of this semantic wave function.

\subsection{Lagrangian Density and Nonlinearity}

We begin by establishing a baseline using the linear Schrödinger equation. While LLMs 
are inherently nonlinear systems due to their neural network architectures and training processes, analyzing 
the linear case first provides a valuable point of comparison for understanding the effects of incorporating 
nonlinearity. This approach allows us to isolate and understand the specific contributions of nonlinear 
terms in shaping the dynamics of the semantic wave function.

To introduce nonlinearity, we consider two distinct mechanisms: (1) the addition of a cubic term to the 
wave equation, resulting in a nonlinear Schrödinger equation, and (2) the introduction of a Mexican hat potential.

The nonlinear Schrödinger equation, characterized by a cubic term, is a common starting point for 
studying nonlinear quantum systems. This nonlinearity allows the semantic wave function to interact with itself. 
The strength and nature of this self-interaction is determined by the coupling constant, $\gamma$. A positive 
coupling constant ($\gamma > 0$) can be interpreted as semantic self-reinforcement, where the presence of a 
particular semantic meaning enhances itself. Conversely, a negative coupling constant ($\gamma < 0$) represents 
semantic self-inhibition, where the presence of a semantic meaning suppresses its own presence. These 
interpretations allow us to relate the parameter $\gamma$ to the semantic coherence and diversity of the LLM.

The second approach to introducing nonlinearity involves the Mexican hat potential. This potential, already 
discussed in earlier sections, provides an alternative mechanism for capturing nonlinear effects and for the 
spontaneous symmetry breaking that can happen within the LLM.

Based on these considerations, we propose the following Lagrangian density for the semantic wave function,

\begin{equation}
    \mathcal{L} = 
    \mathcal{L}_S +\mathcal{L}_N 
    =\frac{i\hbar}{2}\left(\psi^*\frac{\partial \psi}{\partial t}-\psi\frac{\partial \psi^*}{\partial t}\right)
    -\frac{\hbar^2}{2m}\sum_{i=1}^{N} \left |\frac{\partial \psi}{\partial x_i} \right |^2
    +\mathcal{L_{N}}.
\end{equation}

\noindent
Here, $\psi(t, \vec{x})$ represents the semantic wave function,  
where $N$ is the dimension of the embedded space with spatial coordinates $\vec{x}$. Setting $\hbar = 1$ 
and $m = 1$ implies that we are measuring energy and momentum in dimensionless units, which simplifies the 
mathematical treatment but requires careful consideration when interpreting the physical meaning of the parameters. 
The Lagrangian $\mathcal{L}_S $ epresents the linear Schrödinger equation and corresponds to the propagation 
of a classical free particle, while the term $\mathcal{L_{N}}$ encompasses the nonlinear components.

For the nonlinear Schrödinger equation, the nonlinearity has the form

\begin{equation}
    \mathcal{L_{N}} = \mathcal{L_{NS}} =
    -\frac{\gamma}{2}|\psi|^4,
\end{equation}

\noindent
where $\gamma$ is the coupling constant mentioned prior, and it determines the strength of the nonlinear 
self-interaction term.

Alternatively, for the Mexican hat potential, the nonlinearity is described by

\begin{equation}
    \mathcal{L_{N}} = \mathcal{L_{MH}} =
     \mu^2 |\psi|^2 - 2\lambda |\psi|^4.
\end{equation}

\noindent
Both $\mu^2$ and $\lambda$ are positive parameters. This potential, $V(\psi) = - \mu^2 |\psi|^2 + 2\lambda |\psi|^4$, 
exhibits minima at $|\psi| = v = \sqrt{\mu^2/(2\lambda)}$, representing stable semantic states.

\subsection{Introduction of Gauge Field and U(1) Symmetry}

To ensure the stability of our model and prevent the arbitrary creation or destruction of semantic information, 
we impose a U(1) symmetry. This symmetry guarantees that the total "semantic charge" is conserved. In quantum field 
theories, U(1) symmetry is associated with the conservation of electric charge, and we draw an analogy to the 
conservation of semantic meaning.

To implement the U(1) symmetry, we introduce a gauge field, $A_{\mu}(t, \vec{x})$, where $\mu = 0, 1, ..., N$. 
We also use the notation

\begin{equation}
  x_{\mu}\hspace*{0.5cm}\mu = 0,1,..,N \hspace*{0.2cm}{\rm with}
  \hspace*{0.2cm}x_0=t,\hspace*{0.3cm}(x_1,x_2,...,x_{N})=(x_i) = \vec{x},
\end{equation}

\noindent
i.e.

\begin{equation}
  x_{\mu} = (x_0,x_i) = (t,\vec{x}), \hspace*{0.3cm}i=1,2,...,N.
\end{equation}

\noindent
$A_0$ is the scalar potential, and $A_i$ is the vector potential. 
The gauge transformations for the fields are

\begin{align}
  \psi(x_{\mu}) &\rightarrow \psi'(x_{\mu}) = e^{ig\varphi(x_{\mu})} \psi(x_{\mu}), \phantom{\frac{1}[1]}\\
  \psi^*(x_{\mu}) &\rightarrow \psi'^*(x_{\mu}) = e^{-ig\varphi(x_{\mu})} \psi^*(x_{\mu}), \phantom{\frac{1}[1]}\\
  A_{\mu}(x_{\mu}) &\rightarrow A'_{\mu}(x_{\mu}) = A_{\mu}(x_{\mu}) + \frac{1}{g}\partial_{\mu}\varphi(x_{\mu}).
\end{align}

\noindent
Here, $g$ is the charge associated with the field $\psi$ and $\varphi(x_{\mu})$ is a gauge parameter. We have also used
the notation $\partial_{\mu} = \partial/\partial x_{\mu}$.
We replace the ordinary derivatives with covariant derivatives,

\begin{equation}
  D_{\mu}  = \partial_{\mu} - igA_{\mu}.
\end{equation}

\noindent
We set $g=1$. 
The gauge-invariant Lagrangian is then

\begin{equation}
    \mathcal{L} = -\frac{1}{4}F_{\mu\nu} F^{\mu\nu}+ \frac{i}{2}(\psi^* D_0 \psi - \psi (D_0 \psi)^*)
    - \frac{1}{2} |D_i \psi|^2 +\mathcal{L_{N}},
\end{equation}

\noindent
where

\begin{equation}
    F_{\mu\nu} = \partial_\mu A_\nu - \partial_\nu A_\mu.
\end{equation}

\noindent
Repeated indices are summed except when otherwise indicated. Expanding the Lagrangian, we get

\begin{equation}
  \mathcal{L} = -\frac{1}{4}F_{\mu\nu} F^{\mu\nu}+ \frac{i}{2} ( \psi^* \partial_0 \psi
  - \psi \partial_0\psi^* ) + A_0 |\psi|^2
  - \frac{1}{2}  |\partial_i \psi|^2 - \frac{1}{2}  A_i^2 |\psi|^2
  - \frac{i}{2}  A_i (\psi^* \partial_i \psi - \psi \partial_i \psi^*) +\mathcal{L_{N}}.
  \label{eq_lagrangian}
\end{equation}

\noindent
This is the Lagrangian describing a wave propagating in a nonlinear semantic space with U(1) symmetry.

\subsection{Gauge Fixing (Coulomb Gauge)}

The gauge freedom in the Lagrangian leads to redundant degrees of freedom, which need to be removed to 
obtain a well-defined path integral. To do this, we choose the Coulomb gauge,

\begin{equation}
 \partial_i A_i = 0.
\end{equation}

\noindent
This choice simplifies calculations and has a physical interpretation in our context, as we will discuss later. 
To implement the Coulomb gauge, we use the Faddeev-Popov procedure. We can define number "1",

\begin{equation}
1 = \int \mathcal{D}[\alpha] \det(-\partial_i^2) \delta\left[\partial_i A_i - \alpha\right],
\end{equation}

\noindent
where $\partial_i^2$ is the Laplacian in $N$ dimensions, $\delta$ is a delta function and $\alpha$ is some function. Then, we insert this into the 
path integral. We also introduce a Lagrange multiplier field $\lambda$ to enforce the gauge condition. 
After integrating out the Lagrange multiplier, we have a generating functional,

\begin{equation}
\begin{split}
Z = \int  \mathcal{D}[\psi] \mathcal{D}[\psi^*] \mathcal{D}[A]  \delta\left[\partial_i A_i \right] 
 \exp\left(i S[\psi, \psi^*, A] \right),
\end{split}
\end{equation}

\noindent
where the action is

\begin{equation}
\begin{split}
S[\psi, \psi^*, A] = \int dt \int d^{N}x \left( - \frac{1}{4}  F_{\mu\nu} F^{\mu\nu} +\frac{i}{2} (
    \psi^* \partial_0 \psi
- \psi \partial_0 \psi^* ) + A_0 |\psi|^2 - \frac{1}{2}  |\partial_i \psi|^2  \right.  \\
 \left. - \frac{1}{2}  A_i^2 |\psi|^2 - \frac{i}{2}  A_i (\psi^* \partial_i \psi - \psi \partial_i \psi^*)
 +\mathcal{L_{N}} \right).
\end{split}
\end{equation}

\noindent
The determinant $\det(-\partial_i^2)$ is a number and can be absorbed into the path integral.

\subsection{Mean-Field Approximation}

Directly integrating out the gauge fields $A_\mu$ to obtain an effective action that depends 
solely on $\psi$ and $\psi^*$ poses significant challenges. 
Therefore, we implement the mean-field approximation, a common technique to simplify such calculations. 
This simplification, while enabling analytical progress, neglects quantum fluctuations and correlations,
potentially overlooking subtle but important aspects of the semantic dynamics. The validity of the mean-field 
approximation relies on the assumption that fluctuations around the mean field are small. This assumption is more 
likely to hold when the system is strongly interacting or when the number of degrees of freedom is large. 
In the context of LLMs, this could correspond to situations where the semantic space is highly structured or when 
the LLM has a large number of parameters.

In the mean-field approximation, we decompose the quantum fields into their average (classical) values and 
fluctuations around these averages. Now $\psi$ is a complex field and needs some attention, but we may express it 
and the gauge field $A_\mu$ as

\begin{align}
\psi(x_{\mu}) &= \langle \psi(x_{\mu}) \rangle + \delta \psi(x_{\mu}) = \bar{\psi}(x_{\mu}) + \delta \psi(x_{\mu}), \\
A_\mu(x_{\mu}) &= \langle A_\mu(x_{\mu}) \rangle + \delta A_\mu(x_{\mu}) = \bar{A}_\mu(x_{\mu}) + \delta A_\mu(x_{\mu}).
\end{align}

\noindent
Here, $\bar{\psi}(x_{\mu})$ and $\bar{A}_\mu(x_{\mu})$ denote the mean fields, while $\delta \psi(x_{\mu})$ 
and $\delta A_\mu(x_{\mu})$ represent the corresponding fluctuations. We invoke the assumption that these 
fluctuations are small in magnitude compared to the mean fields, which is a central tenet of the mean-field approximation. 
In the context of LLMs, interpreting the mean field $\bar{\psi}(x_{\mu})$ as the mean field, it provides how 
common states happen in the embedding space.

Next, we treat the term $|\psi|^2$ as a background "semantic charge" density. Within this approximation, we can 
solve for the component $A_0$ of the gauge field in terms of $\psi$ and $\psi^*$. Employing the Coulomb gauge, 
the equation of motion for $A_0$ is derived by applying the Euler-Lagrange equation. Concretely, 
we suppose that we're evaluating the amount of this action with respect to a "ground" (classical field) level of activity 
that measures whether or not we've hallucinated.

The equation of motion for $A_0$ is then obtained by varying the action with respect to $A_0$,

\begin{equation}
\frac{\delta S}{\delta A_0} = 0 \implies |\psi(x_{\mu})|^2 - \partial_j^2 A_0(x_{\mu}) = 0,
\end{equation}

\noindent
from which we get

\begin{equation}
-\partial_j^2 A_0(x_{\mu}) = - \frac{\partial^2 A_0(x_{\mu})}{\partial x_i^2} = |\psi(x_{\mu})|^2.
\end{equation}

\noindent
Therefore,

\begin{equation}
A_0(x_{\mu}) = \int d^{N} x' \, G(x_{\mu}, x_{\mu}') \, |\psi(x_{\mu}')|^2,
\end{equation}

\noindent
where $G(x_{\mu}, x_{\mu}')$ is the Green's function for the Laplacian operator in $N$ dimensions satisfying the equation

\begin{equation}
   -\partial_i^2 G(x_{\mu}, x_{\mu}') = \delta^N(x_{\mu} - x_{\mu}').
\end{equation}

\noindent
The Green's function 
represents the potential at point $x_{\mu}$ due to a unit charge at point $x_{\mu}'$. For the specific case of $N = 3$, 
the Green's function is given by $G(x_{\mu}, x_{\mu}') = -1/(4\pi |x_{\mu}-x_{\mu}'|)$. For the general case of $N$ 
dimensions, 
the Green's function is expressed as

\begin{equation}
G(x_{\mu},x_{\mu}') = -\frac{\Gamma\left(N/2\right)}{2(N-2)\pi^{N/2} |x_{\mu}-x_{\mu}'|^{N-2}}.
\end{equation}

\noindent
Using these results we get
the effective action

\begin{equation}
    \begin{split}
    S_{eff} = \int dt \int d^{N}x \left[\frac{i}{2} \left( \psi^* \partial_0 \psi 
    - \psi \partial_0 \psi^* \right) - \frac{1}{2}  |\partial_i \psi|^2 
    - \frac{1}{2}  A_i^2 |\psi|^2  
    - \frac{i}{2}  A_i (\psi^* \partial_i \psi - \psi \partial_i \psi^*) + \mathcal{L_{N}}\right] \\
    -\frac{1}{2} \int dt \int d^{N}x \int d^{N}x' G(x_{\mu}, x_{\mu}') |\psi(x_{\mu})|^2 |\psi(x_{\mu}')|^2.
    \end{split}
\end{equation}

\subsection{Interpretation of Terms}

The effective action provides a framework for understanding how semantic meanings interact and evolve
within the LLM embedding space. We can analyze the key terms in the effective action to gain insights into these 
dynamics. The gauge fields, particularly the vector potential $A_i$, play a crucial role in shaping this semantic 
landscape, acting as directional forces that guide the flow of semantic information. While the precise nature of 
these forces remains speculative, we can explore some concrete examples of how they might manifest within LLMs.

The kinetic energy term, proportional to $|\partial_t\psi_i|^2$, relates to the smoothness or coherence of semantic 
meaning.
A smaller gradient implies a more uniform or consistent meaning across the semantic space, while larger gradients
may indicate ambiguity or rapid shifts in meaning. This term is expressed as

\begin{equation}
    \frac{i}{2}\left(\psi^* \partial_0 \psi - \psi \partial_0 \psi^*\right).
\end{equation}

\noindent
The term governs the temporal evolution of the semantic wave function. It dictates how the semantic
meaning changes over time, with the rate of change being influenced by the other terms in the Lagrangian.

The gradient term for $\psi$, given by

\begin{equation}
    -\frac{1}{2}(\partial_i \psi^*) (\partial_i \psi),
\end{equation}

\noindent
penalizes rapid changes in the semantic wave function across the semantic space. This term encourages semantic
smoothness and coherence. A large gradient would correspond to abrupt shifts in meaning, which are generally
disfavored in coherent text generation. This term can be related to the concept of "semantic distance" -
the further apart two points are in the semantic space, the more "costly" it is to transition between them.

The interaction term between $\psi$ and $A_i$ is

\begin{equation}
    -\frac{1}{2} \left[ i A_i (\psi^* \partial_i \psi - \psi \partial_i \psi^*) + A_i^2 \psi^* \psi \right],
\end{equation}

\noindent
which couples the semantic wave function $\psi$ to the vector potential $A_i$.
The semantic current interaction,

\begin{equation}
    i A_i (\psi^* \partial_i \psi - \psi \partial_i \psi^*) = i A_i J_i,
\end{equation}

\noindent
where $J_i = \psi^* \partial_i \psi - \psi \partial_i \psi^*$ is the "semantic current." This term represents the
interaction between the vector potential and the flow of semantic information. It describes how the gauge field
influences the movement of semantic meaning within the LLM.

One concrete example of how the vector potential could influence this flow is through its analogy 
to attention mechanisms in Transformer networks. Consider the sentence, "The dog chased the ball, 
but it was too fast." When processing "it," the vector potential (represented by the 
attention weights) would strongly connect "it" to "ball," indicating that "it" refers to the ball. 
This connection guides the flow of semantic information, ensuring that the LLM correctly understands 
the pronoun reference. Without this directional influence, the LLM might incorrectly 
associate "it" with "dog." In this sense, the attention 
weights can be seen as components of the vector potential, directing the semantic current ($J_i$) towards 
the relevant parts of the input sequence.

The semantic density interaction is

\begin{equation}
    A_i^2 \psi^* \psi.
\end{equation}

\noindent
This term is proportional to the "semantic density" $\psi^* \psi$ and represents the direct interaction between
the vector potential and the presence of a semantic concept. It describes how the gauge field responds to the
presence of semantic meaning. This could be related to how the LLM adjusts its internal state based on the 
semantic content of the input.

The kinetic term for $A_i$,

\begin{equation}
    -\frac{1}{4}  (\partial_i A_j - \partial_j A_i)(\partial_i A_j - \partial_j A_i),
\end{equation}

\noindent
represents the kinetic energy of the vector potential. This term governs the dynamics of the gauge field itself.
It ensures that the gauge field is not static but evolves in response to the semantic wave function.

The Coulomb interaction term is given by

\begin{equation}
    -\frac{1}{2} \psi^*(x_{\mu}) \psi(x_{\mu}) \int d^{N}x' G(x_{\mu}, x_{\mu}') \psi^*(x_{\mu}') \psi(x_{\mu}').
\end{equation}

\noindent
This term represents the effective, scaled interaction between semantic charge densities, mediated by the 
underlying gauge field. The factor of -1/2 reflects the dynamics of this mediating field. The 
term $|\psi(x_{\mu})|^2$ can be interpreted as the semantic charge density at point $x_{\mu}$ in the 
semantic space, with higher values indicating a stronger presence of that particular meaning at 
that location. The distance $|x_{\mu} - x_{\mu}'|$ represents the semantic distance between points 
$x_{\mu}$ and $x_{\mu}'$ in the semantic space, with smaller distances implying a closer semantic 
relationship. This term suggests that words or phrases with similar semantic charges and located 
close to each other in the semantic space will experience an attractive or repulsive force, 
depending on the sign of $G(x_{\mu}, x_{\mu}')$. The -1/2 factor scales the magnitude of this interaction. 
This long-range interaction could be related to the ability of LLMs to capture dependencies between 
distant parts of a text, although the strength of these dependencies is now scaled down. The Green's 
function $G(x_{\mu}, x_{\mu}')$ determines the strength and form of this interaction.

In summary, the effective action provides a rich framework for understanding the dynamics of semantic meaning in LLMs.
Each term in the action corresponds to a specific aspect of semantic interaction and evolution. By analyzing these terms,
we can gain insights into how LLMs process and generate language. It is important to emphasize that these 
interpretations are speculative, and further research is needed to validate these connections. However, 
they provide concrete examples of how the vector potential could play a role in guiding the flow of 
semantic information within LLMs.

\subsection{Understanding the Cubic Nonlinearity}

The cubic nonlinearity in the nonlinear Schrödinger equation (NLSE) introduces a term that depends on the 
intensity of the wave function, $|\psi|^2$. This implies that the wave function interacts with itself, and the 
strength and nature of this interaction are determined by the coupling constant $\gamma$. The sign and 
magnitude of $\gamma$ significantly influence the behavior of the semantic wave function and can be 
interpreted in several ways.

If $\gamma > 0$, the nonlinearity represents semantic self-reinforcement. In this scenario, the presence of 
a particular semantic meaning reinforces itself, and the stronger the initial presence of a meaning, the more 
it tends to amplify itself. This could be analogous to priming, where a concept already activated in the LLM's 
representation becomes easier to activate further. It might also relate to confirmation bias, where the LLM tends 
to favor interpretations consistent with its prior beliefs or knowledge, or to semantic coherence, where the 
LLM tries to maintain a consistent semantic representation. Mathematically, this would lead to a self-focusing effect, 
where the wave function tends to concentrate in regions where it is already strong.

Conversely, if $\gamma < 0$, the nonlinearity represents semantic self-inhibition. In this case, 
the presence of a particular semantic meaning inhibits itself, and the stronger the initial presence of a meaning, 
the more it tends to suppress itself. This could be analogous to attention mechanisms, where the LLM focuses 
its attention on the most relevant concepts and suppresses irrelevant ones, or to inhibition of return, 
where the LLM avoids repeating the same concepts or ideas too frequently. It could also relate to semantic diversity, 
where the LLM tries to explore different aspects of a topic and avoid getting stuck in a narrow focus. 
Mathematically, this would lead to a self-defocusing effect, where the wave function tends to spread out and 
avoid concentrating in any one region.

It is also possible that the sign and magnitude of $\gamma$ depend on the context. In some contexts, a concept 
might reinforce itself ($\gamma > 0$), while in other contexts, it might inhibit itself ($\gamma < 0$). 
This context-dependent semantic interaction could be analogous to polysemy, where the same word can have different 
meanings in different contexts, or to irony and sarcasm, where the intended meaning of a statement can be 
the opposite of its literal meaning. It could also relate to topic shifts, where the LLM can change its focus 
depending on the conversation or the task. Modeling this would require a more complex model where $\gamma$ is 
a function of the input prompt or the internal state of the LLM.

The cubic nonlinearity affects the interaction between different semantic components, represented by 
different wave functions. To model this, one would need to consider a system of coupled NLSEs, one for each 
semantic component. 
The nonlinearity would then involve terms like $|\psi_i|^2 \psi_j$, which represent the influence of the i-th 
component on the j-th component. This could be analogous to semantic networks, where concepts are connected 
to each other in a network, and the activation of one concept can influence the activation of other concepts, 
or to argument structure, where the meaning of a sentence depends on the relationships between its different parts. 
This would lead to a much more complex model, but it could capture more nuanced semantic interactions.

In summary, the cubic nonlinearity in the NLSE model offers a way to capture interactions within the semantic 
space of LLMs. The specific interpretation of this nonlinearity, and particularly the sign and magnitude of the 
coupling constant $\gamma$, depends on the specific semantic phenomena being modeled. Further research is needed 
to determine the most appropriate interpretation and to develop more sophisticated models that can capture 
the complexities of semantic meaning and interaction.

\subsection{Spontaneous Symmetry Breaking and Semantic Bias}

The key difference in Mexican hat potential to the cubic nonlinearity is that the Mexican hat potential 
provides a mechanism for spontaneous symmetry breaking. The semantic field $\psi$ will tend to settle into 
one of the minima of the potential, breaking the symmetry of the Lagrangian. This could be interpreted as 
the LLM "choosing" a particular interpretation or perspective, even in the absence of specific input.

The interplay between the kinetic energy term, the Mexican hat potential, and the Coulomb interaction could 
lead to the emergence of complex semantic structures in the LLM embedding space. The Mexican hat potential 
provides a stable foundation for these structures, while the Coulomb interaction allows for long-range 
interactions between different semantic concepts.

The spontaneous symmetry breaking inherent in the Mexican hat potential has profound implications for the 
interpretation of LLMs. The choice of a particular vacuum state $v = |v|\exp(i\theta)$ represents a fundamental 
bias in the LLM's semantic representation. The magnitude $|v| = \sqrt{\mu^2/(2\lambda)}$ determines the overall 
strength of this bias, while the phase $\theta$ determines the specific direction of the bias in the complex 
semantic space.

This bias could manifest in various ways, such as a preference for certain types of information, a tendency 
to adopt certain viewpoints, or a predisposition to generate certain types of text. It is important to note 
that this bias is not necessarily a negative thing. It could be a reflection of the LLM's training data or 
its intended purpose. However, it is important to be aware of this bias and to understand how it might 
influence the LLM's behavior.

Furthermore, the fluctuations around the vacuum state can be interpreted as the specific semantic 
content of the LLM's output. These fluctuations are influenced by the input prompt, the LLM's internal state, 
and the interactions between different semantic concepts.

In summary, the path integral formulation with the Mexican hat potential provides a powerful framework 
for understanding the dynamics of semantic meaning in LLMs. The spontaneous symmetry breaking inherent 
in this model leads to the emergence of a fundamental semantic bias, which influences the LLM's behavior and shapes 
its output.

\subsection{Detailed Calculation of Effective Action with Integrated Out $A_i$}

This section provides a detailed calculation of the effective action obtained by integrating out the vector 
potential $A_i$ to quadratic order, under the weak coupling approximation and in the Coulomb gauge. 
The weak coupling approximation assumes that the interaction between the semantic wave function and the 
gauge field is small, which may not always be valid in LLMs.

We begin with the action that includes the vector potential $A_i$,

\begin{equation}
    S[A] = \int dt \int d^{N}x \left\{ -\frac{1}{4}  (\partial_i A_j - \partial_j A_i)^2 
    - \frac{1}{2}  A_i^2 \psi^* \psi  + \frac{1}{2i}  A_i (\psi^* \partial_i \psi - \psi \partial_i \psi^*) \right\}.
\end{equation}

\noindent
The first term involves the field strength tensor $F_{ij} = \partial_i A_j - \partial_j A_i$. 
We integrate this term by parts,

\begin{equation}
    -\frac{1}{4} \int dt \int d^{N}x  (\partial_i A_j - \partial_j A_i)^2 
    = \frac{1}{2} \int dt \int d^{N}x  A_j (-\partial_i^2 A_j + \partial_i \partial_j A_i).
\end{equation}

\noindent
We then apply the Coulomb gauge condition, $\partial_i A_i = 0$, which eliminates the second term,

\begin{equation}
    \frac{1}{2} \int dt \int d^{N}x  A_j (-\partial_i^2 A_j + \partial_i \partial_j A_i) 
    = \frac{1}{2} \int dt \int d^{N}x  A_i (-\partial_j^2 A_i),
\end{equation}

\noindent
where $\partial_i^2$ is the Laplacian.
Substituting this back into the action, we get

\begin{equation}
    S[A] = \int dt \int d^{N}x \left\{ \frac{1}{2} A_i (-\partial_j^2 A_i) - \frac{1}{2}  A_i^2 \psi^* \psi 
     + \frac{1}{2i} A_i (\psi^* \partial_i \psi - \psi \partial_i \psi^*) \right\}.
\end{equation}

\noindent
We rewrite the action in a form suitable for Gaussian integration,

\begin{equation}
    S[A] = \frac{1}{2} \int dt \int d^{N}x  \left\{ A_i (-\partial_j^2 - \psi^* \psi) A_i 
    + \frac{1}{i} A_i (\psi^* \partial_i \psi - \psi \partial_i \psi^*) \right\}.
\end{equation}

\noindent
Now we can identify the kernel $K_{ij}(x_{\mu},y_{\mu})$ and the source $J_i(x_{\mu})$,

\begin{eqnarray}
        K_{ij}(x_{\mu},y_{\mu}) &=& (-\partial^2_j 
        - \psi^*(x_{\mu}) \psi(x_{\mu})) \delta_{ij} \delta^N(x_{\mu}-y_{\mu}), \\
        J_i(x_{\mu}) &=& \frac{1}{i} (\psi^*(x_{\mu}) \partial_i \psi(x_{\mu}) 
        - \psi(x_{\mu}) \partial_i \psi^*(x_{\mu})).
\end{eqnarray}

\noindent
The path integral over $A_i$ is now a Gaussian integral. The general formula for a Gaussian integral is

\begin{equation}
    \int DA \exp\left\{{\frac{i}{2} \int dx dy A(x) K(x,y) A(y) + i \int dx J(x) A(x)}\right\} 
    = \text{constant} \times \exp\left\{{-\frac{i}{2} \int dx dy J(x) K^{-1}(x,y) J(y)}\right\}.
\end{equation}

\noindent
Applying this formula to our action, we get

\begin{equation}
    \int DA \exp\left\{iS[A]\right\} \propto 
    \exp\left\{-\frac{i}{2} \int dt \int d^{N}x d^{N}y J_i(x) K^{-1}_{ij}(x_{\mu},y_{\mu}) J_j(y_{\mu})\right\}.
\end{equation}

\noindent
This means that the effective action, after integrating out $A_i$, is

\begin{equation}
    S_{eff} = S_0 - \frac{1}{2} \int dt \int d^{N}x d^{N}y J_i(x_{\mu}) 
    K^{-1}_{ij}(x_{\mu},y_{\mu}) J_j(y_{\mu}) + S_N,
\end{equation}

\noindent
where $S_0$ is the action for the $\psi$ field without the $A_i$ field.
The inverse of the kernel, $K^{-1}_{ij}(x_{\mu},y_{\mu})$, is called the propagator. 
It satisfies the following equation

\begin{equation}
    \int dt \int d^{N}z K_{ik}(x_{\mu},z_{\mu}) K^{-1}_{kj}(z_{\mu},y_{\mu}) = \delta_{ij} \delta^N(x_{\mu}-y_{\mu}).
\end{equation}

\noindent
Substituting the expression for $K_{ij}(x_{\mu},y_{\mu})$, we get

\begin{equation}
    \int dt \int d^{N}z (-\partial_j^2 - \psi^*(x_{\mu}) \psi(x_{\mu})) \delta_{ik} 
    \delta^N(x_{\mu}-z_{\mu}) K^{-1}_{kj}(z_{\mu},y_{\mu}) = \delta_{ij} \delta^N(x_{\mu}-y_{\mu}).
\end{equation}

\noindent
This simplifies to

\begin{equation}
    (-\partial_j^2 - \psi^*(x_{\mu}) \psi(x_{\mu})) K^{-1}_{ij}(x_{\mu},y_{\mu}) 
    = \delta_{ij} \delta^N(x_{\mu}-y_{\mu}).
\end{equation}

\noindent
To make progress, we make the weak coupling approximation. This means that we assume that the interaction 
between the $\psi$ field and the $A_i$ field is weak. This allows us to neglect the $\psi^*(x_{\mu}) \psi(x_{\mu})$ 
term in the equation for the propagator. The validity of this approximation depends on the specific values 
of the parameters in the Lagrangian and the properties of the semantic wave function. In LLMs, this 
approximation might be reasonable if the semantic charge density is relatively low or if the gauge coupling is weak.

The inverse Kernel satisfies the following equation

\begin{equation}
    (-\partial_j^2) K^{-1}_{ij}(x_{\mu},y_{\mu}) = \delta_{ij} \delta^N(x_{\mu}-y_{\mu}).
\end{equation}

\noindent
The solution to the simplified equation is

\begin{equation}
    K^{-1}_{ij}(x_{\mu},y_{\mu}) = \delta_{ij} G(x_{\mu},y_{\mu}),
\end{equation}

\noindent
where $G(x_{\mu},y_{\mu})$ is the Green's function for the Laplacian in $N$ dimensions. 
The Green's function satisfies the equation

\begin{equation}
    (-\partial_j^2) G(x_{\mu},y_{\mu}) = \delta^N(x_{\mu}-y_{\mu}).
\end{equation}

\noindent
The solution for $G(x_{\mu}, x_{\mu}')$ is N-dependent:
\begin{equation}
    G(x_{\mu},x_{\mu}') = -\frac{\Gamma\left(N/2\right)}{2(N-2)\pi^{N/2} |x_{\mu}-x_{\mu}'|^{N-2}} \quad (N \neq 2).
\end{equation}

\noindent
The Green's function represents the influence of a semantic "charge" at point $y_{\mu}$ on the potential at point $x_{\mu}$. 
It effectively describes how semantic information propagates through the semantic space.
Substituting the expression for the propagator into the effective action, we get

\begin{equation}
    S_{eff} = S_0 - \frac{1}{2} \int dt \int d^{N}x d^{N}y J_i(x_{\mu}) G(x_{\mu},y_{\mu}) J_i(y_{\mu}) +S_N.
\end{equation}

\noindent
The effective Lagrangian is then

\begin{equation}
\begin{split}
    \mathcal{L}_{eff} &= \frac{i}{2}\left(\psi^* \partial_0 \psi - \psi \partial_0 \psi^*\right)
    - \frac{1}{2}\sum (\partial_i \psi^*) (\partial_i \psi)  \\
    &- \frac{1}{2}  \int d^{N}y \left( \psi^*(x_{\mu}) \partial_i \psi(x_{\mu})
    - \psi(x_{\mu}) \partial_i \psi^*(x_{\mu}) \right) G(x_{\mu},y_{\mu}) \left( \psi^*(y_{\mu}) \partial_i \psi(y_{\mu}) 
    - \psi(y_{\mu}) \partial_i \psi^*(y_{\mu})
    \right) + \mathcal{L}_{N}.
\end{split}
\label{eq_eff_lagrangian}
\end{equation}

\noindent
This effective Lagrangian describes the dynamics of the $\psi$ field, taking into account the effects of the 
integrated-out $A_i$ field to quadratic order in the weak coupling approximation. The next to last term represents a 
non-local interaction between the $\psi$ field at different points in space, mediated by the Green's 
function $G(x_{\mu},y_{\mu})$. This non-local interaction implies that the semantic meaning at one point in the 
semantic space can directly influence the semantic meaning at other points, even if they are far apart. 
This could be related to the ability of LLMs to capture long-range dependencies in text.

The effective action, derived from this Lagrangian, provides a powerful tool for analyzing the behavior of LLMs. 
By studying the solutions to the equations of motion derived from this action, we can gain insights into 
how LLMs process and generate language. However, it is important to remember that this model is based on several 
simplifying assumptions, and its predictions should be interpreted with caution. Further research is needed 
to validate this model and to explore its implications for our understanding of LLMs.

\subsection{Interpretation of the Effective Lagrangian: Impact of Approximations}

The effective Lagrangian, derived after integrating out the gauge field $A_i$ and employing 
the weak coupling and mean-field approximations, provides a modified perspective on the 
semantic dynamics within LLMs compared to the initial Lagrangian. It is crucial to 
understand how these approximations influence the interpretation of the terms and the overall model.

In the original Lagrangian (Equation \ref{eq_lagrangian}), the interaction 
between the semantic wave function $\psi$ and the gauge field $A_i$ was explicit 
and local. Terms involving $A_i$ directly coupled the semantic current and density 
to the vector potential at the same spatial point, representing a direct and immediate 
influence of the gauge field on the semantic wave function.

However, the effective Lagrangian (Equation \ref{eq_eff_lagrangian}), 
obtained after integrating out $A_i$ and applying the weak coupling approximation, 
exhibits a key difference: the emergence of a non-local interaction term. This term, 
involving the Green's function $G(x, y)$, couples the semantic current at point $x$ 
to the semantic current at point $y$, integrated over the entire semantic space.

This non-locality has several important implications. The original Lagrangian 
primarily captured local interactions. In contrast, the effective Lagrangian 
explicitly incorporates long-range dependencies between different regions of 
the semantic space. This suggests that the semantic meaning at one location can 
directly influence the semantic meaning at distant locations, mediated by the 
Green's function. This is particularly relevant in the context of LLMs, which are 
known for their ability to capture long-range dependencies in text. The non-local 
interaction, mediated by the Green's function, may be viewed as analogous to the 
self-attention mechanism in Transformer architectures \cite{transformer}, which
enables the model to attend to different parts of the input sequence, irrespective of 
their proximity.

Furthermore, the integration of $A_i$ effectively generates an "effective potential" 
governing the interaction between semantic elements. This potential is encoded in the 
Green's function $G(x,y)$, which dictates the strength and form of the interaction 
between points $x$ and $y$. The shape of this potential is determined by the 
underlying Laplacian operator and the dimensionality of the semantic space.

The approximations employed to derive the effective Lagrangian inevitably lead 
to a loss of microscopic details. By integrating out $A_i$, we have averaged over 
its fluctuations and correlations. Consequently, the effective Lagrangian only 
captures the average effect of the gauge field on the semantic wave function. 
The weak coupling approximation further simplifies the interaction, potentially 
neglecting higher-order effects.

The mean-field approximation replaces the quantum fields with their average values, 
neglecting quantum fluctuations. This implies that the effective Lagrangian describes 
the behavior of the average semantic wave function, rather than the behavior of 
individual semantic elements. The validity of this approximation hinges on the 
assumption that fluctuations are small compared to the mean field, which may not 
always hold in LLMs.

In the original Lagrangian, the semantic current $J_i$ directly interacted with 
the vector potential $A_i$, representing a local flow of semantic information. 
In the effective Lagrangian, the semantic current at one point influences the 
semantic current at another point through the Green's function. This suggests 
a more distributed and interconnected flow of semantic information across 
the semantic space.

In summary, the effective Lagrangian provides a simplified but potentially 
insightful framework for understanding the dynamics of semantic meaning in LLMs. 
The approximations made to obtain this Lagrangian introduce both advantages and 
limitations. The non-local interaction term captures long-range dependencies, 
potentially mirroring the function of self-attention, but the loss of microscopic 
details and the mean-field perspective limit the accuracy of the model. 
Despite these limitations, the effective Lagrangian offers valuable insights 
into how LLMs process and generate language, warranting further investigation 
and empirical validation.

\subsection{Physical Interpretation of the Semantic Field}

In the context of LLMs, our quantum-like model allows for a speculative, yet potentially insightful, 
interpretation of the gauge fields as representing different aspects of semantic meaning. The scalar potential, 
$\phi$ (or $A_0$), can be interpreted as reflecting the overall semantic context or "energy" within the LLM, 
influencing the probability of activating semantically aligned words. This captures the static contextual 
bias or "semantic atmosphere" surrounding a word or phrase, reflecting the overall influence of the 
surrounding concepts. A high scalar potential might correspond to a state where the LLM is highly engaged 
with a particular topic, leading to a higher probability of generating words related to that topic. 
Conversely, a low scalar potential might indicate a state of low engagement or uncertainty. It could 
also bias the LLM towards certain concepts learned during training, influencing token probabilities 
and potentially relating to the magnitude of hidden state vectors. For example, when processing a 
prompt about "climate change," the scalar potential would be high in regions of the semantic space 
associated with environmental issues, scientific terms, and political debates, increasing the likelihood 
of generating relevant terms.

The vector potential, $A$, on the other hand, can be interpreted as representing the semantic 
relationships or "forces" between words and phrases, guiding the flow of semantic information and 
shaping local interactions. This captures the dynamic flow of semantic information, reflecting 
shifts in topic, argument structure, and narrative flow, and possibly indicating the relationships 
and dependencies between different concepts. The vector potential could be related to the attention 
weights in Transformer networks, encoding relationships and directing the flow of semantic 
information. It could also encode the argument structure of verbs, specifying semantic roles, 
or guide topic shifts and narrative flow. 

Together, $\phi$ and $A$ constitute a "semantic field" that governs the behavior of the semantic 
wave function, $\psi$. While our current model simplifies the analysis by primarily focusing on 
the scalar potential and often neglecting the vector potential, future research should explore 
the potential benefits of incorporating the vector potential to achieve a more complete 
nderstanding of the dynamic interplay of semantic meanings within LLMs. However, it is crucial 
to acknowledge the limitations of this analogy, particularly the gauge dependence of $A$ and 
$\phi$, and to emphasize the need for empirical validation to support these theoretical 
interpretations. The semantic field is likely an emergent property of the LLM's complex neural 
network architecture, and its true nature remains a subject for further investigation. These 
interpretations are speculative, and future work should focus on developing experiments to test 
these hypotheses, such as analyzing attention weights or measuring the influence of perturbations 
on the semantic space.

\section{Assumptions and Limitations}

The quantum-like model presented in this paper relies on several simplifying 
assumptions, which are important to acknowledge and discuss. These assumptions 
allow us to develop a tractable mathematical framework, but they also introduce 
limitations that could affect the validity of our conclusions and the scope 
of our interpretations.

\subsection{Completeness of Vocabulary}

One key assumption is that the LLM's finite vocabulary forms an approximately 
complete basis for representing semantic information. This assumption allows 
us to treat the semantic space as a discrete space built upon a finite set 
of basis states, simplifying the mathematical formalism and enabling the 
analogy to quantum systems with discrete energy levels. However, it is 
crucial to recognize that the vocabulary is not truly complete. New words 
and phrases are constantly being created, and the LLM's vocabulary may not 
capture all of the nuances of human language, especially in specialized 
domains or rapidly evolving cultural contexts. Furthermore, the use of 
sub-word tokenization, while helpful for handling out-of-vocabulary words, 
can create semantic units that are not explicitly present as single 
entries in the vocabulary, potentially blurring the lines between 
discrete basis states.

This limitation could affect the model's ability to accurately represent 
semantic concepts that are not well-represented in the vocabulary. For example, 
the model might struggle to understand slang terms, technical jargon, or newly 
coined words. It might also lead to inaccuracies when dealing with rare or 
specialized language, or when encountering subtle semantic distinctions 
that are not explicitly encoded in the vocabulary. The assumption of 
vocabulary completeness also neglects the dynamic nature of language, 
where word meanings evolve and new concepts emerge over time.

\subsection{Linearity of Operations on the Semantic Space 
(Given Nonlinear Training)}

In several parts of our analysis, we make assumptions that implicitly 
treat operations on the semantic space as linear. It's crucial to acknowledge 
that this is a simplification, given that the LLM's training process is 
inherently nonlinear, and the resulting embedding space reflects this 
nonlinearity. While we perform linear operations on the learned embeddings, 
the embeddings themselves are the product of a highly nonlinear process. 
This "inherited nonlinearity" means that our linear operations are still 
influenced by the underlying nonlinear structure of the embedding space, 
but they may not be able to fully capture its richness. These linearity 
assumptions manifest in several key aspects of the model:

\begin{enumerate}
    \item Plane Wave Approximations: The approximation of semantic waves as 
    plane waves, solutions to linear wave equations, simplifies wave behavior 
    and neglects potential distortions or interactions that would arise in 
    a nonlinear medium.
    \item Superposition of Semantic States: The representation of a semantic 
    state as a linear superposition of basis states, $\ket{\alpha} = \sum  
    c_i\ket{\psi_i}$, assumes that the meaning of a complex phrase is simply 
    the sum of the meanings of its individual components, weighted by coefficients. 
    This neglects the possibility of emergent meanings arising from the interaction 
    of words.
    \item Linear Schrödinger Equation (Initial Formulation): The initial 
    use of the linear Schrödinger equation to model semantic wave propagation 
    assumes that the semantic wave function evolves linearly, without 
    self-interaction or interaction with other semantic components.
    \item Complex-Valued Similarity Measure: The complex-valued similarity 
    measure, $S_T(\text{text1},\text{text2}) = \sum c_{1i}^* c_{2i}$, 
    calculates similarity by summing the products of complex coefficients, 
    a linear operation. This assumes that overall similarity is a sum of 
    component similarities, neglecting nonlinear relationships between components.
    \item The derivation of the Green's function relies on solving a simplified, 
    linearized differential equation (the Laplacian). This simplification 
    arises from the combined effect of the mean-field approximation and the 
    weak coupling approximation. The mean-field approximation linearizes the 
    problem around the average fields, neglecting fluctuations. The weak 
    coupling approximation further simplifies the equation by dropping terms 
    that couple the semantic wave function and the gauge fields. While the 
    full problem is inherently nonlinear due to terms in the Lagrangian that 
    couple $\psi$ and $A_\mu$, these approximations allow us to obtain an 
    analytic expression for the Green's function, which would otherwise be 
    intractable. This linearized Green's function then describes the propagation 
    of the average semantic potential, neglecting nonlinear effects and fluctuations.
 \end{enumerate}

The reliance on linear operations limits the model's ability to capture complex 
semantic phenomena. For example, consider the sentence "That's sick!" In modern 
slang, "sick" can mean "amazing" or "excellent," a meaning that is not a linear 
combination of the dictionary definition of "sick." Similarly, the meaning of a 
metaphor like "Time is a thief" cannot be derived by simply adding the meanings 
of "time" and "thief." Irony and sarcasm also rely on nonlinear inversions of 
meaning. The model's reliance on cosine similarity, a linear measure, further 
reinforces this limitation.

\subsection{Mean-Field Approximation}

When applying the path integral formalism, we employ the mean-field approximation, 
which assumes that the quantum fields can be approximated by their average values, 
neglecting quantum fluctuations. This approximation greatly simplifies the 
calculations and allows us to obtain an effective action that depends only 
on the average semantic wave function, providing a tractable framework for 
analyzing the dynamics of the semantic space. However, the mean-field approximation 
neglects important fluctuations and correlations, which could play a significant 
role in the dynamics of the semantic space, particularly at smaller scales 
or in highly dynamic contexts.

Consequently, the model's ability to capture subtle semantic effects and 
emergent phenomena, such as the spontaneous emergence of new meanings 
or the rapid shifts in topic that can occur in natural language, could 
be affected. It might also 
lead to inaccuracies when dealing with highly dynamic or unpredictable language, 
where fluctuations play a more prominent role. The mean-field approximation 
also assumes a degree of homogeneity in the semantic space, which may not 
always be valid.

\subsection{Weak Coupling Approximation}

In deriving the effective action, we also employ the weak coupling approximation, 
which assumes that the interaction between the semantic wave function and the 
gauge field is weak. This assumption simplifies the calculation of the effective 
action and allows us to obtain an analytical solution, providing a more manageable 
mathematical framework. However, the interaction between the semantic wave 
function and the gauge field might be strong in some cases, particularly when 
dealing with highly charged or polarized semantic concepts, such as those a
ssociated with strong emotions or deeply held beliefs.

This may have implications for the model's ability to accurately capture 
the influence of the gauge field on the semantic wave function in these situations. 
It might also lead to inaccuracies when dealing with highly emotional or 
persuasive language, where the gauge field plays a more prominent role in 
shaping the semantic landscape. The weak coupling approximation also implies 
that the semantic charge density is relatively low, which may not always be 
the case in LLMs that have learned to represent complex and nuanced semantic 
relationships.

\subsection{Static Potential Landscapes}

When modeling semantic ambiguity using potential landscapes (e.g., 
double-well potential, Mexican hat potential), we often assume that these 
landscapes are static and do not change over time. This assumption simplifies 
the analysis and allows us to focus on the equilibrium states of the semantic 
wave function, providing a more tractable framework for understanding how 
LLMs handle multiple meanings. However, the potential landscapes are likely 
to be dynamic and to change over time in response to the input prompt, the 
LLM's internal state, and the interactions between different semantic concepts. 
The context in which a word is used can significantly alter its meaning and 
the corresponding potential landscape.

The assumption of a static potential landscape could limit the model's ability 
to capture the dynamic evolution of semantic meaning and the influence of 
context on word meanings. 
It might also lead to inaccuracies when dealing with language that is highly 
dependent on context or that involves rapid shifts in meaning. The assumption 
of static potential landscapes also neglects the learning process, where the 
potential landscapes themselves are shaped by the LLM's training data.

\subsection{Specific Gauge Choice (Coulomb Gauge)}

In applying the path integral formalism, we have chosen the Coulomb gauge, 
which simplifies calculations and has a physical interpretation in terms of 
semantic charge conservation. However, the Coulomb gauge is not the only 
possible gauge choice, and other gauge choices might provide different 
insights into the behavior of the semantic space. The physical interpretation 
of the gauge fields is also gauge-dependent, meaning that the meaning we 
ascribe to the scalar and vector potentials depends on the gauge we choose.

The choice of the Coulomb gauge could influence the interpretation of the 
results and limit the generality of the model. While the Coulomb gauge 
provides a convenient 
framework for understanding semantic charge conservation, it might not be 
the most natural or appropriate choice for all situations. Other gauge 
choices might reveal different aspects of the semantic landscape and provide 
alternative interpretations of the gauge fields.

\subsection{Simplified Representation of LLM Architecture}

It is also crucial to acknowledge that our model provides a highly 
simplified representation of the complex neural network architecture of 
LLMs. We have abstracted away many of the details of the Transformer architecture, 
including the multi-head attention mechanism, the feedforward networks, 
and the residual connections. While these simplifications allow us to develop 
a tractable mathematical framework, they also limit the model's ability to 
capture the full complexity of LLM behavior. The model does not explicitly 
account for the role of different layers in the network, the flow of information 
between layers, or the specific activation functions used in the neurons.

The simplified nature of the model could affect the model's ability to accurately 
predict the behavior of LLMs in all situations. It might also limit the model's ability 
to provide insights into the specific mechanisms that drive LLM performance. 
The model should be seen as a high-level abstraction that captures some of the 
key principles underlying LLM behavior, rather than a detailed simulation.

\section{Hallucinations and Semantic Charge Conservation: A Question for Future Investigation}

This section explores a potential, albeit speculative, link between the constraints imposed by 
our model and the phenomenon of hallucinations in LLMs. We frame this connection as a question for 
future investigation, aiming to provide a new perspective on the origins of untruthful or nonsensical 
text generation. Within our model, the imposition of the Coulomb gauge can be interpreted as 
enforcing a form of semantic charge conservation. Analogous to charge conservation in electromagnetism, 
we hypothesize that the total semantic meaning within the LLM's representation 
(represented by the integral of $|\psi|^2$ over the semantic space) remains approximately constant. 
This constraint reflects the finite vocabulary and the underlying structure learned during training. 
The LLM, in essence, redistributes semantic meaning across its vocabulary to represent different 
concepts and relationships, but the total amount of meaning remains bounded. A key question arises: 
does the stability of this semantic charge conservation correlate with the generation of truthful 
and coherent text? We propose that deviations from this idealized conservation, potentially arising 
from noise, incomplete modeling of contextual influences, or inherent limitations in the LLM's 
representation, might be indicative of, or even directly contribute to, instances of hallucination.

Several mechanisms could lead to deviations from semantic charge conservation. The LLM might fail 
to fully account for the context in which a word or phrase is used, leading to an inappropriate 
redistribution of semantic charge. For example, if the LLM misinterprets a negation, it might 
assign semantic charge to the wrong concept, leading to a factual error. Random noise in the 
LLM's internal state could disrupt the delicate balance of semantic charge, leading to the 
creation of spurious meanings or the amplification of irrelevant concepts. If the LLM has 
overfit its training data, it might simply memorize specific facts or patterns without truly 
understanding the underlying semantic relationships. In this case, it might generate text 
that is factually correct but semantically incoherent, effectively violating semantic charge 
conservation. The complex, non-linear interactions within the deep neural network 
architecture of LLMs could lead to emergent phenomena that are not captured by our simplified model.

Alternatively, even within a system that nominally conserves semantic charge, inherent noise 
and fluctuations in the model's dynamics could lead to the transient emergence of spurious 
or "hallucinatory" meanings. These fluctuations could manifest as virtual particles within 
the semantic space, momentarily borrowing energy to create fleeting, nonsensical semantic 
constructs. The creation and annihilation of these virtual particles could, in effect, lead 
to a temporary violation of semantic charge conservation at a local level, even if the 
overall charge remains approximately conserved. This perspective suggests that techniques 
used to regularize LLMs and prevent overfitting might be seen as analogous to mechanisms 
that suppress the creation of virtual particles in a quantum field theory, thereby maintaining 
the integrity of the semantic space. These techniques effectively dampen fluctuations and 
encourage the LLM to learn more generalizable semantic representations.

While these connections are currently speculative, they suggest that our framework offers 
a new perspective for investigating the relationship between model constraints, semantic 
stability, and the generation of truthful text.

\section{Discussion}

This paper presents a quantum-like model offering a new perspective on the inner workings 
of Large Language Models (LLMs). By drawing analogies to concepts from quantum mechanics, 
such as superposition, interference, potential landscapes, and path integrals, we have 
developed a framework for understanding how semantic meaning is represented and 
processed within these complex systems.

The application of quantum mechanical concepts to the analysis of LLM embedding spaces 
is a relatively unexplored area, offering the potential for new insights. The model provides 
intuitive explanations for several observed phenomena in LLMs, such as the probabilistic 
nature of their outputs, their ability to handle semantic ambiguity, and the long-range 
dependencies between words and phrases. The use of mathematical tools from quantum mechanics, 
such as the Schrödinger equation and the path integral formalism, provides a more rigorous 
framework for analyzing LLM behavior than purely descriptive approaches, potentially leading 
to new tools and techniques for analyzing, interpreting, and manipulating LLMs based on 
quantum mechanical principles.

However, it is crucial to acknowledge that LLMs, despite their impressive capabilities, 
remain essentially "black boxes." Their internal workings are complex and opaque, making 
it difficult to understand precisely how they process and generate language. The quantum-like 
model presented in this paper is not intended to be a complete or definitive explanation of 
LLM behavior. Rather, it is an attempt to provide a more interpretable framework for 
understanding their behavior, offering a new set of tools and concepts for analyzing their 
internal representations and dynamics. By drawing analogies to quantum mechanics, we hope to 
shed light on the emergent properties of LLMs and to inspire new avenues for research.

The model is based on analogies and interpretations, rather than direct empirical evidence, 
and it is not yet clear whether LLM embedding spaces actually behave in the way predicted. 
It is likely an oversimplification of the complex dynamics of semantic meaning, as many 
factors known to influence LLM behavior, such as the training data, the model architecture, 
and the specific task, are not explicitly accounted for. The physical interpretation of the 
vector potential $A$ and the scalar potential $\phi$ in the context of LLMs requires further 
clarification. The use of the mean-field approximation further simplifies the model, 
potentially neglecting important fluctuations and correlations, and the model has not 
yet been rigorously tested against empirical data from LLMs.

Key results include demonstrating that the probabilistic nature of LLM outputs can be 
understood in terms of a quantum two-level system, arguing that extending the embedding 
space to the complex domain is necessary to capture semantic interference effects, 
developing a quantum mechanical formulation of semantic representation using a "semantic wave function," 
showing how potential landscapes can be used to model semantic ambiguity, introducing a new, 
complex-valued similarity measure, presenting a path integral formalism for modeling LLM behavior, 
discussing the implications of Coulomb gauge fixing, and speculating on the connection between 
deviations from semantic charge conservation and hallucinations.

Several broad research directions emerge from this work. Further investigation into the concept 
of semantic charge, its conservation properties, and its relationship to LLM behavior is 
warranted. This includes developing methods for measuring semantic charge, exploring the 
mechanisms that lead to deviations from conservation, and investigating the connection between 
semantic charge conservation and the generation of truthful and coherent text. Exploring the 
potential of quantum computing to enhance LLM capabilities is a promising area for future 
research. If the semantic space exhibits a truly quantized structure, quantum algorithms could 
offer advantages in tasks such as efficient semantic search, where quantum search algorithms, 
like Grover's algorithm, could potentially accelerate the process of finding the closest 
semantic matches for a given query in the embedding space. Quantum simulation techniques 
could be used to model the dynamic evolution of semantic wave functions, potentially 
capturing emergent phenomena and subtle semantic interactions that are difficult to simulate 
classically. Quantum machine learning algorithms could be explored for training LLMs, 
potentially leading to models with improved performance, new capabilities, or more efficient 
training processes. 

Leveraging the extensive body of knowledge surrounding path integrals 
in physics offers a rich set of tools for analyzing LLM dynamics. 
Adapting techniques from quantum field theory and high-energy physics, such as perturbation theory, 
renormalization group methods, and the development of Feynman diagrams, could provide 
new insights into the interactions between semantic concepts and the emergence of complex 
semantic structures. Developing "classical LLM Feynman diagrams" to visualize and calculate 
effective actions could offer a powerful framework for understanding and predicting LLM behavior. 
In addition, insights from nonlinear optics are directly applicable to this endeavor.

We could also reconsider our approach to the inherent probabilistic behavior of LLMs. 
While current efforts largely focus on mitigating specific inaccuracies like "hallucinations," 
our quantum-like model suggests a broader perspective. If LLMs operate within a fundamentally 
probabilistic semantic space governed by quantum-like principles, then this inherent 
uncertainty is not merely a source of errors, but a fundamental characteristic affecting all 
aspects of their operation. Instead of solely trying to suppress specific manifestations of 
this probabilistic behavior, perhaps we can learn to harness it. Just as quantum microscopes 
exploit quantum fluctuations to achieve resolutions beyond classical limits, could we 
leverage the inherent uncertainty of LLMs to generate new ideas, explore creative possibilities, 
or uncover hidden relationships within data? This paradigm shift, from error correction to 
opportunistic exploitation of inherent probabilistic behavior, could unlock new and unexpected 
applications for LLMs.

\end{document}